\documentclass[10pt,twocolumn,letterpaper]{article}

\usepackage[utf8]{inputenc} %
\usepackage[T1]{fontenc}    %
\usepackage{url}            %
\usepackage{booktabs}       %
\usepackage{amsfonts}       %
\usepackage{nicefrac}       %
\usepackage{microtype}      %

\usepackage{3dv}
\usepackage{times}
\usepackage{epsfig}
\usepackage{graphicx}
\usepackage{amsmath}
\usepackage{amssymb}
\usepackage{mathrsfs}

 \usepackage{relsize}

\usepackage{sidecap}

\makeatletter
\@namedef{ver@everyshi.sty}{}
\makeatother
\usepackage{tikz}

\usetikzlibrary{decorations.pathreplacing,calligraphy}
\usetikzlibrary{shapes,arrows,chains,positioning}
\usetikzlibrary{shapes.geometric}
\usetikzlibrary[calc]

\usepackage{transparent}

\usepackage[pagebackref=true,breaklinks=true,letterpaper=true,colorlinks,bookmarks=false]{hyperref}

\usepackage{xcolor}
\hypersetup{
	colorlinks,
	citecolor={red!50!black},
	urlcolor={blue!80!black}
}
\tikzset{
	ncbar angle/.initial=90,
	ncbar/.style={
		to path=(\tikztostart)
		-- ($(\tikztostart)!#1!\pgfkeysvalueof{/tikz/ncbar angle}:(\tikztotarget)$)
		-- ($(\tikztotarget)!($(\tikztostart)!#1!\pgfkeysvalueof{/tikz/ncbar angle}:(\tikztotarget)$)!\pgfkeysvalueof{/tikz/ncbar angle}:(\tikztostart)$)
		-- (\tikztotarget)
	},
	ncbar/.default=0.5cm,
}
\tikzset{round left paren/.style={ncbar=0.5cm,out=120,in=-120}}
\tikzset{round right paren/.style={ncbar=0.5cm,out=60,in=-60}}

\newcommand{\fis}{f_{I \rightarrow S}}
\newcommand{\fsi}{f_{S \rightarrow I}}

\threedvfinalcopy %

\def\threedvPaperID{345} %

\begin{document}

\title{Cycle-Consistent Generative Rendering for 2D-3D Modality Translation}

\author{Tristan Aumentado-Armstrong\\
{\tt\small tristan.a@partner.samsung.com}
\and
Alex Levinshtein\\
{\tt\small alex.lev@samsung.com}
\and
Stavros Tsogkas\\
{\tt\small stavros.t@samsung.com}
\and
Konstantinos G.\ Derpanis\\
{\tt\small k.derpanis@samsung.com}
\and
Allan D.\ Jepson\\
{\tt\small allan.jepson@samsung.com}
\and
\hspace*{2in}
Samsung AI Centre Toronto
\hspace*{2in}
}

\maketitle
\widowpenalty=0
\clubpenalty=0

\begin{abstract}
For humans, visual understanding is inherently generative: given a 3D shape, we can postulate how it would look in the world; given a 2D image, we can infer the 3D structure that likely gave rise to it. We can thus translate between the 2D visual and 3D structural modalities of a given object. In the context of computer vision, this corresponds to a learnable module that serves two purposes: (i) generate a realistic rendering of a 3D object (shape-to-image translation) and (ii) infer a realistic 3D shape from an image (image-to-shape translation). In this paper, we learn such a module while being conscious of the difficulties in obtaining large paired 2D-3D datasets. By leveraging generative domain translation methods, we are able to define a learning algorithm that requires only weak supervision, with unpaired data. The resulting model is not only able to perform 3D shape, pose, and texture inference from 2D images, but can also generate novel textured 3D shapes and renders, similar to a graphics pipeline. More specifically, our method (i) infers an explicit 3D mesh representation, (ii) utilizes example shapes to regularize inference, (iii) requires only an image mask (no keypoints or camera extrinsics), and (iv) has generative capabilities. While prior work explores subsets of these properties, their combination is novel. We demonstrate the utility of our learned representation, as well as its performance on image generation and unpaired 3D shape inference tasks.
\end{abstract}

\begin{figure}
	\centering
	\begin{tikzpicture}[node distance=3cm,main node/.style={}]
	\node[main node,align=center](centerlabel) {
		\phantom{Unpaired}\\
		\phantom{Data}
	};
	\node[main node,align=center,xshift=1mm] at (centerlabel) (centerlabeltext) {
		Unpaired\\
		Data
	};
	\node[main node,left=1mm of centerlabel](imgs) {
		\includegraphics[width=0.15\textwidth]{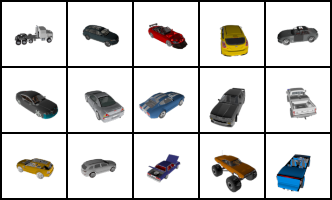}
	};
	\node[main node,right=1mm of centerlabel](shapes) {
		\includegraphics[width=0.19\textwidth,trim={0cm 8cm 0cm 11cm},clip]{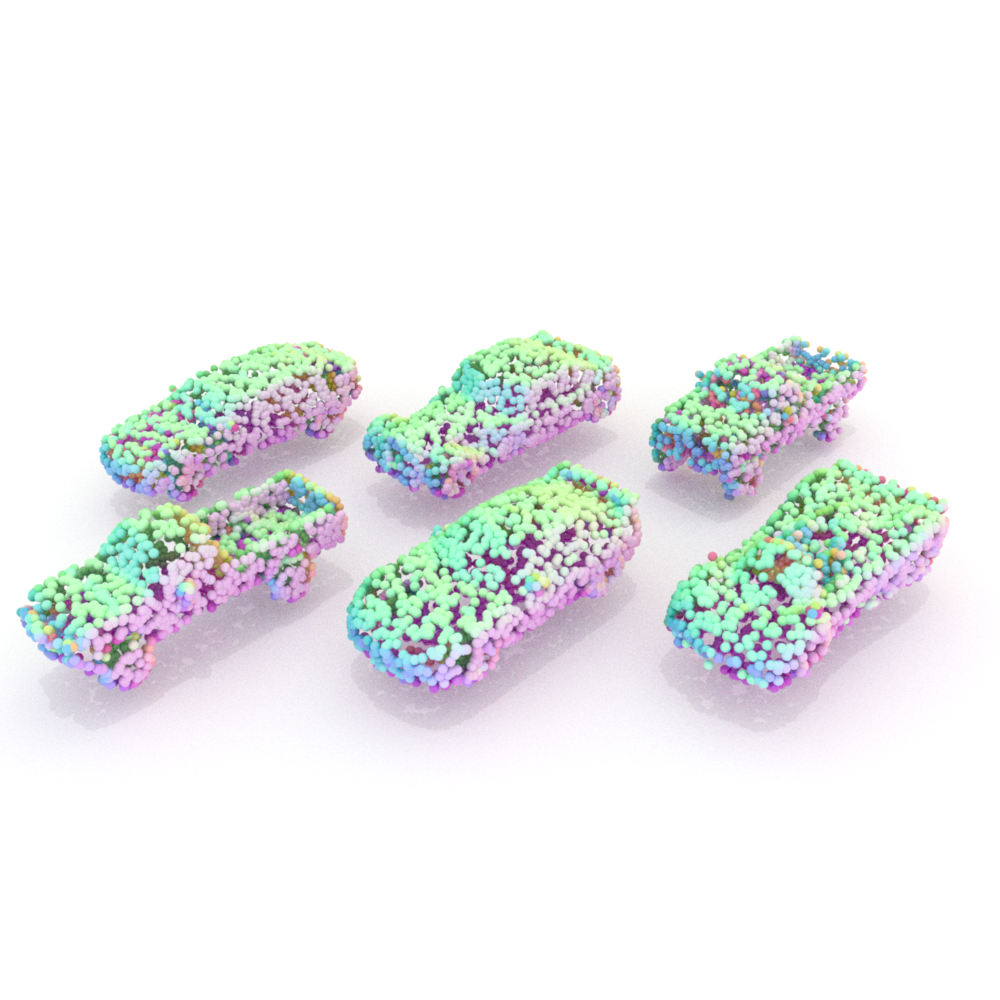}
	};
	\node[main node, below = 5.53mm of imgs,xshift=-8mm,draw,black](i2simg){
		\includegraphics[width=0.04\textwidth,trim={0cm 0.0cm 0cm 0.0cm},clip]{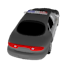}
	};
	
	\node[main node, right = 38mm of i2simg,xshift=-1mm,black](i2sshape){
	};
	\node[main node, right=0mm of i2sshape,xshift=-4mm] (blah){
	};

	\node[main node, right = 38mm of i2simg,xshift=-1mm,black](PC){
		\includegraphics[width=0.057\textwidth,trim={0cm 0.1cm 0cm 0.5cm},clip]{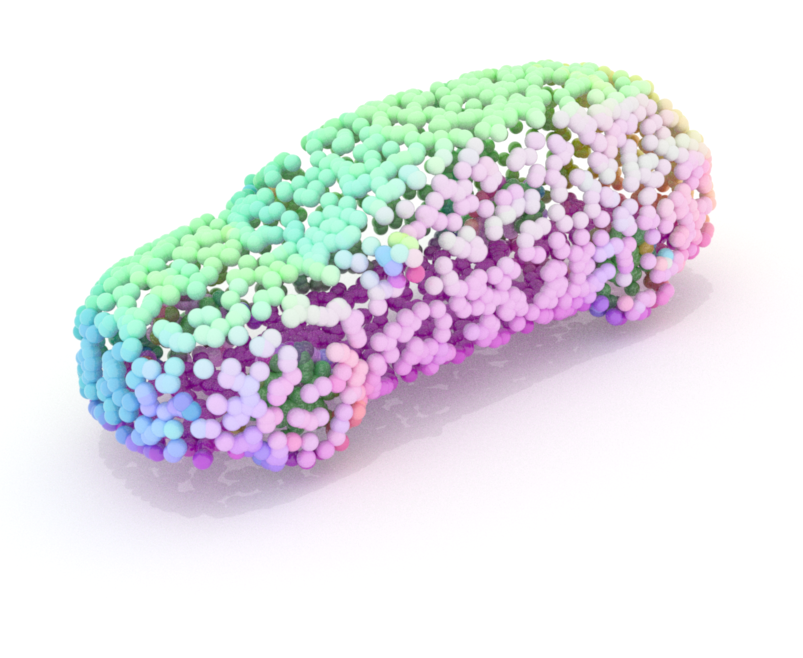}
	};
	\node[main node, right = 1mm of PC,xshift=-4mm](M){
		\includegraphics[width=0.072\textwidth]{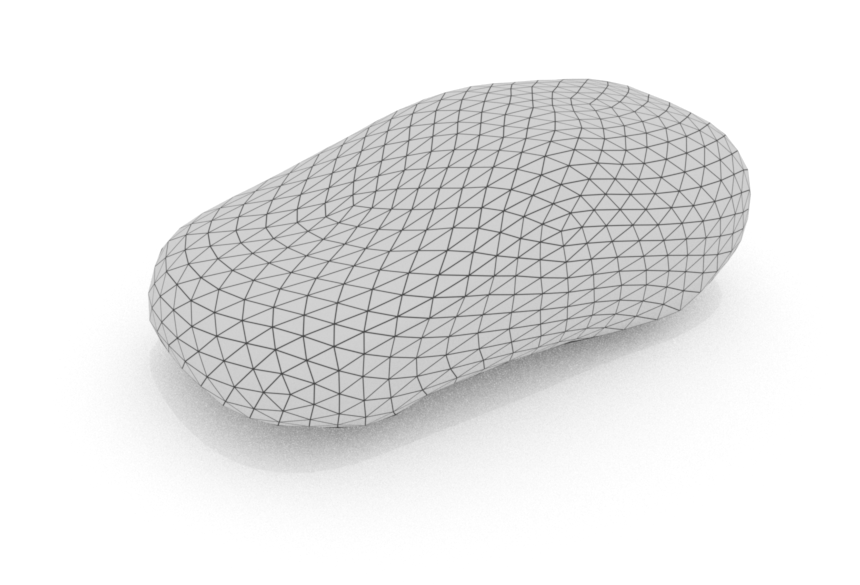}
	};
	\node[main node, right = 1mm of M,xshift=-2mm,yshift=0.5mm](inftex){
		\includegraphics[width=0.033\textwidth]{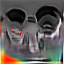}
	};
	\node[main node,xshift=-3mm,yshift=1mm] at (PC.south east) (a){};
	\node[main node,xshift=5mm,yshift=1mm] at (M.north west) (b){};
	
	\path[every node/.style={anchor=south}]
	(a) edge [] node [midway,above,align=center] 
	{} (b);

	\node[main node, left=0.01cm of PC,xshift=0mm,yshift=-4mm,rotate=300] (frust)
	{\includegraphics[width=.04\textwidth]{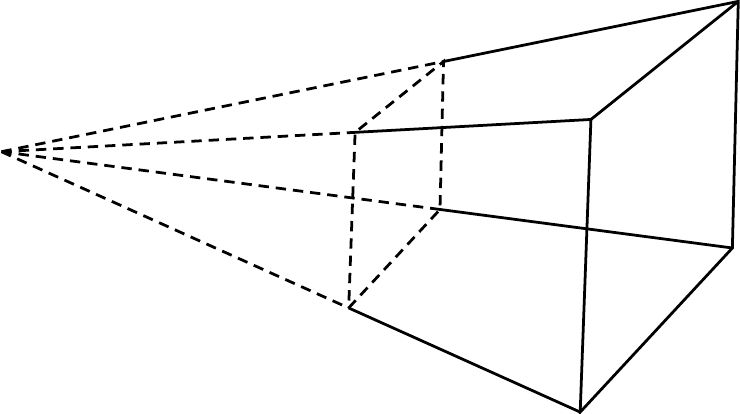}};
	
	\node[below=1mm of frust,align=center,xshift=5.0mm,yshift=-1.9mm] (poselabel) {  Pose\phantom{p}};
	
	\node[right=1mm of poselabel,align=center,xshift=0.2mm,yshift=0mm] (shapelabel) {  Shape};
	
	\node[right=1mm of shapelabel,align=center,xshift=1.5mm,yshift=0mm] (texlabel) { Texture\phantom{p}};
	
	\node[left=1mm of poselabel,align=center,xshift=-27.5mm,yshift=0mm] (texlabel) { Image\phantom{p}};

	\node[main node, right=1mm of i2simg.east,xshift=-2mm] (helper){};
	
	\node[xshift=2mm,yshift=-2mm] at (i2simg.east) (anchorleftlow) {  };
	\node[xshift=2mm,yshift=2mm] at (i2simg.east) (anchorlefthigh) {  };
	\node[xshift=-5mm,yshift=-2mm] at (PC.west) (anchorrightlow) {  };
	\node[xshift=-5mm,yshift=2mm] at (PC.west) (anchorrighthigh) {  };
	
	\path[every node/.style={anchor=south}]
	(anchorrighthigh) edge [->,>=stealth'] node [midway,above,align=center, pos=0.5] 
	{$f_{S\rightarrow I}$} (anchorlefthigh);
	
	\path[every node/.style={anchor=south}]
	(anchorleftlow) edge [->,>=stealth'] node [midway,below,align=center] 
	{$f_{I\rightarrow S}$} (anchorrightlow);
	
	\draw [black] (0.7,-2.4) to [round left paren  ] (0.7,-1.4);
	\draw [black] (4.1,-2.4) to [round right paren] (4.1,-1.4);
	\node [] at (1.25,-2.3) (comma1) {,};
	\node [] at (3.35,-2.3) (comma2) {,};

	\end{tikzpicture}
	\caption{
		 Depiction of our method for learning %
		 the
		 2D-3D
		 \textit{ modality translation} functions, $\fsi$ and $\fis$.
		Given unpaired collections of images and untextured 3D shapes (upper inset), 
		we learn to generate images by rendering shapes with plausible textures and poses ($\fsi$), and infer 3D shape, pose, and texture from an image ($\fis$), as shown in the lower inset.
		Our method takes as input an oriented point cloud in the shape-to-image direction, and infers a mesh representation in the image-to-shape direction.	
	}
	\label{fig:overview}
\end{figure}
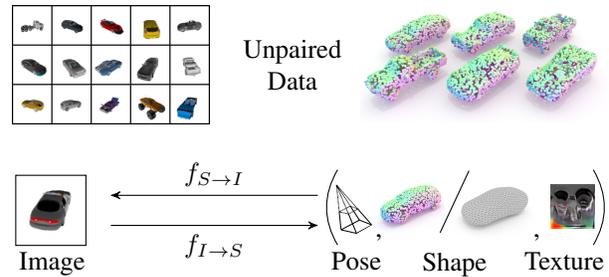

\section{Introduction}
\label{sec:intro}

A natural approach to visual inference is the 
``analysis by synthesis'' paradigm \cite{yuille2006vision}, 
which postulates that cognitive processing proceeds 
in a generative fashion
\cite{yildirim2015efficient}.	
This viewpoint can be referred to as the ``inverse graphics'' characterization 
of computer vision \cite{kulkarni2015deep,yao20183d,tung2017adversarial}.
We expect algorithms of this type to extract a ``graphics code'' from a given image, 
which not only explains the input by reconstructing it, 
but also constitutes an interpretable and manipulable latent representation 
of the image. 
Since the underlying world is three dimensional, 
such a representation should include 3D information as well.
However, learning such a model is non-trivial. 
Even within a single category, the appearance, non-rigid shape, 
and pose of objects can vary widely.
Furthermore, self-occlusions in the image make inferring the correct shape difficult, 
without a strong prior on the allowable shapes, even in relatively simple contexts.

From a modern learning perspective, 
the paucity of paired 2D-3D data presents an additional challenge to learning such inverse graphics models.
While image categories can be annotated by humans with relative ease, 
efficiently obtaining and aligning the correct 3D model for an object into an image 
is difficult, even without considering issues like textures and lighting.
On the other hand, 
synthetically generated paired datasets suffer from a domain shift,
necessitating the use of techniques like domain randomization (DR), 
for helping models transfer from
simulated to real settings \cite{tremblay2018training,tobin2017domain}.
Though useful in many scenarios, DR images are still not realistic, 
and the resulting domain mismatch may thus result in inappropriate data for some tasks.
Separately, it is possible to construct data via 
standard generative models on 2D images 
(e.g., adversarial approaches \cite{goodfellow2014generative}),
but these do not have a naturally interpretable 3D latent representation for labelling or other downstream tasks.

In this paper, we present a weakly supervised model for learning from unpaired data, 
which only requires a set of images and untextured 3D shapes \textit{without correspondence}. 
This allows us to leverage existing 3D datasets,
despite a lack of explicit annotations relating shapes to images.
Similar to cycle-consistent adversarial models, 
capable of domain translation between image types \cite{zhu2017unpaired}, 
we present an algorithm for 
translating between two representational \textit{modalities} of objects:
3D shapes
and 
2D images of those shapes 
(see Fig.\ \ref{fig:overview}).
In other words, our model encapsulates both 
the graphics rendering pipeline, 
mapping a 3D shape to an image, 
and 
the computer vision inference pipeline, 
mapping an image to its latent graphics code,
\textit{within a single learning framework}.
The resulting conditional generative model is able to generate novel images, 
by generating a texture for an input shape and rendering the result;
conversely, we can perform single-image 3D reconstruction, 
by explicitly inferring the 3D shape, pose, and appearance 
of an object 
from a given image.
We use surface meshes as our 3D representation, 
as they circumvent problems 
with memory, resolution, manipulation, and rendering.

We enumerate our contributions as follows:
(a) we define a cyclic generative model on unpaired 2D and 3D data, 
without camera calibration meta-information; %
(b) we show how to utilize adversarial methods to impose a strong prior on 
generated and inferred shapes, textures, poses, and renders;
and
(c) we demonstrate the effectiveness of our model on 
image generation,
single-image textured shape inference, 
and representation learning, 
showing that our method can jointly understand the 2D and 3D modalities of an object.

\begin{figure}
	\centering
	\includegraphics[width=0.48\textwidth]{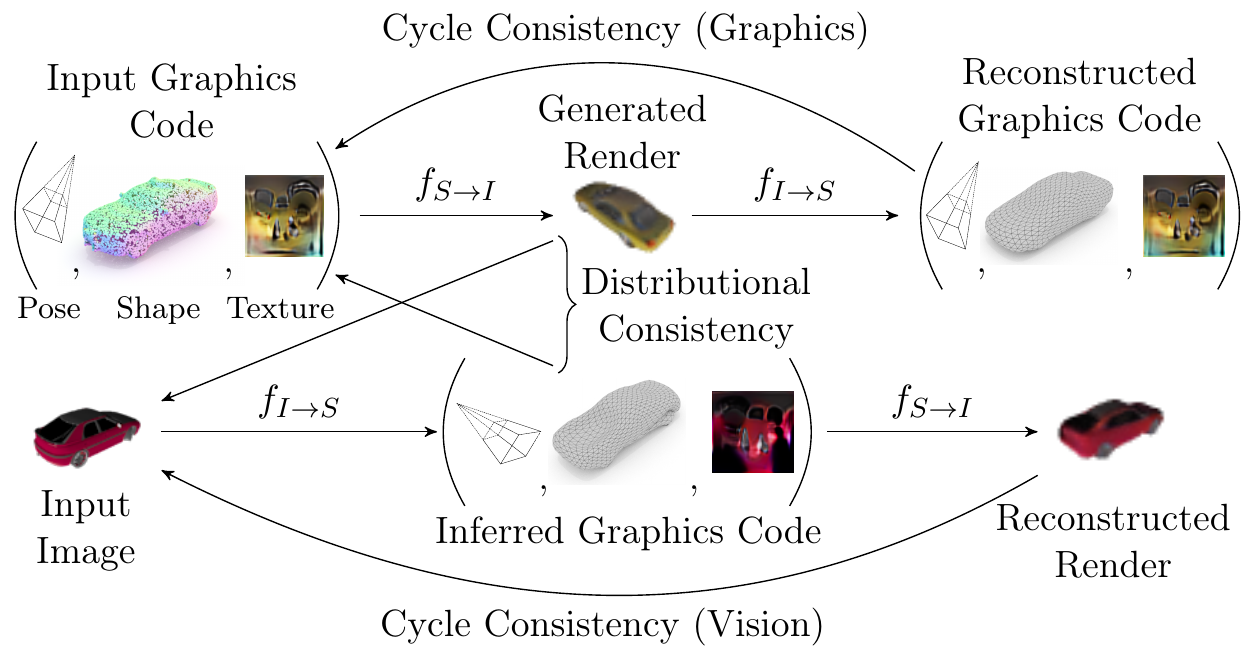}
	\caption{Overview of the cycle-consistent modality translation model. 
		Our two modalities consist of 
		(i) 2D images (assumed to be renders of 3D objects) and 
		(ii) 3D representations, which we show as a \textit{graphics code}, including pose, shape, and texture.
		For shorthand, we refer to this graphics code as simply ``shape''.
		The top pathway depicts the 
		\textit{graphics cycle} (shape-to-image-to-shape), 
		whereas the bottom one shows the 
		\textit{vision cycle} (image-to-shape-to-image), 
		via our modality translation functions $\fis$ and $\fsi$. 
		The primary loss functions are also depicted: 
		(a) the distributional consistency between 
			(i) inferred versus input shapes, and 
			(ii) generated versus real images;
		(b) the cyclic consistency between 
			both the reconstructed and original input shapes and images.
		Note that the graphics cycle starts with a 3D point set, 
			as well as random pose and texture samples, 
			but the inferred and reconstructed codes utilize a mesh 
			(see text for details).
	}
	\label{fig:cycles}
\end{figure}

\section{Related Work}
\label{sec:relwork}

Inferring the 3D structure of an object from a single image 
is among the most studied problems in computer vision,
and continues to receive considerable attention 
\cite{han2019image}.
Several recent works have shown impressive results reconstructing meshes from 2D images 
using 3D supervision 
\cite{wang2018pixel2mesh,gkioxari2019mesh,pan2018residual}.
A different approach is to use
powerful %
parametric models of deformations
 \cite{loper2015smpl,zuffi20173d},
which makes it possible to do highly effective mesh reconstruction from 2D data 
\cite{kanazawa2018end,zuffi2018lions},
but %
requires a pre-existing model of a canonical shape 
and its deformation space.
Another recent trend is the weakening of supervision requirements for 3D inference,
such as via an adversarial prior on the output shapes \cite{jiang2018gal,gwak2017weakly,henzler2018escaping,kanazawa2018end}.
One can also replace 3D supervision with multi-view information \cite{tulsiani2017multi} or keypoints \cite{kanazawa2018learning}.
Tulsiani et al.\ 
\cite{Tulsiani_2018_CVPR} tackle the additional difficult problem of inferring pose 
from images, 
as well as shape,
utilizing only multi-view consistency for weak supervision.
We infer both shape and pose from an image, using only unpaired shape data to form an adversarial prior on the inference process. 

Relatively few machine learning models exist for generative models of textured 3D meshes
\cite{chen2019learning,henderson2020leveraging}.
However, with the advent of differentiable renderers \cite{kato2018neural,li2018differentiable,liu2019soft,chen2019learning}, 
this has become more viable. %
Visual Object Networks \cite{zhu2018visual}
jointly model 3D shape and images in a generative fashion, 
but do not handle full 3D textures.
Raj et al.\ 
\cite{raj2019learning} 
generate 3D textures by first texturing multiple 2.5D views,
and combining the results in a differentiable renderer.
Learned texture fields \cite{oechsle2019texture} 
are able to learn a 3D representation-agnostic generative model of shape textures.
Our model serves as a conditional generative model of textured meshes, %
and also provides a method for inferring them from a single image.

Several recent methods experiment with latent 3D representations (``graphics codes''), 
largely learned from 2D supervision.
In particular, rendering-based generative models
have been some of the earliest models
able to both infer and generate 3D representations
\cite{rezende2016unsupervised}.
Liu et al.\ 
\cite{liu2019learning} investigate 3D reconstruction and generation from 2D images 
using implicit surfaces.
PlatonicGAN \cite{henzler2018escaping} and IG-GAN \cite{lunz2020inverse}
utilize a generative model on 2D images, based on rendering voxels.
HoloGAN \cite{nguyen2019hologan} and BlockGAN \cite{nguyen2020blockgan}
utilize 3D transformations of feature maps to achieve an implicit 3D representation
within a generative model.
Works that utilize learned neural renderers 
(rather than ``hand-designed'' ones)
may gain an advantage in optimization, 
but lose some guarantees, such as correct viewpoint invariance 
\cite{nguyen2018rendernet,nguyen2019hologan}.
3D-SDN \cite{yao20183d} enables 3D-aware image editing 
via an inverse graphics approach,
utilizing renderers and de-renderers for shape and texture
to reconstruct the image.
Other approaches also use rendering-based mesh reconstructions 
in a generative auto-encoding model 
\cite{henderson2019learning,henderson2020leveraging,kanazawa2018learning}.
These latter approaches are most similar to our work, 
except that we infer both pose and texture,
and utilize unpaired 3D data to impose a prior on the reconstructions.

Another closely related vision technique is the use of 3D information 
for inter-image correspondence inference.
Canonical surface mapping learns a mapping 
from 2D image pixels to a 3D template,
implicitly allowing correspondence computation
\cite{kulkarni2019canonical,kulkarni2020articulation,tulsiani2020implicit}.
The recent work by You et al.\ \cite{you2020semantic} consider a 2D-3D-2D cyclic model, 
to estimate corresponding keypoints in images.
Similarly, our model automatically performs correspondence estimation without the need for keypoint data, 
due to the use of a canonical template and a cycle-consistency requirement to match inferred nodal positions to those in the original input shape.

Lastly, our model may be viewed as a form of 2D-3D modality translation.
Such conditional generative modelling techniques impose a cycle-consistency criterion between two domains, and thus provide a powerful framework for domain adaptation \cite{hoffman2017cycada}.
The prototypical example of this family of models is CycleGAN \cite{zhu2017unpaired}, 
but they do not focus on 3D information.
Miyauchi et al.\ \cite{miyauchi2018shape} 
define a cycle-consistent model between an image and its surface normal.
Adversarial Inverse Graphics networks \cite{tung2017adversarial} 
are able to perform 2D-3D lifting with unpaired supervision.
These approaches do not attempt to completely translate 
between the 2D and 3D object representation modalities;
hence, they do not completely recover 3D shape, pose, and texture as we do.
Furthermore, several recent works focus on 
the synthetic-to-real domain translation problem
imposed by 2D image data generated from 3D renderings
\cite{peng2018syn2real,peng2018synthetic}.
Others attempt to induce invariance via domain randomization
\cite{tremblay2018training,tobin2017domain}.
In contrast, our work uses distribution matching losses to enforce
the generated renders to resemble in-domain images in a data-driven fashion.
Finally, we operate on the 3D shape data level (before rendering), 
rather than altering the resulting renders or simply randomizing the textures.

\begin{figure*}
	\centering
	\begin{tikzpicture}[main node/.style={}]
	\node[main node] (inshape)
	{\includegraphics[width=.16025\textwidth]{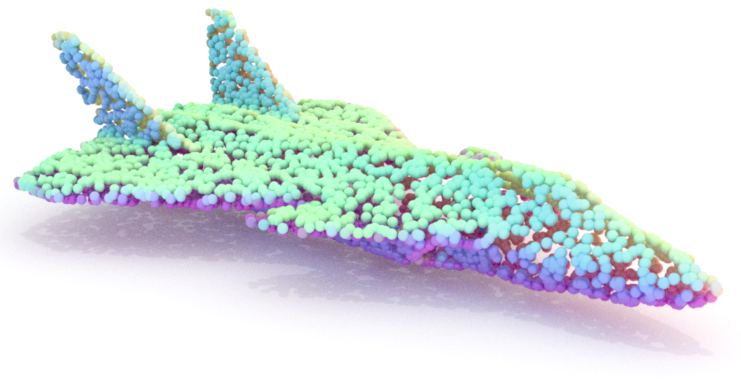}};
	\node[yshift=0.97cm] at (inshape) (inshapeL) 
	{Input Shape};
	\node[yshift=0.65cm,xshift=0.0cm] at (inshape) (inshapeLr) 
	{ \large $P$ };	
	\node[main node, right=1.075cm of inshape,circle,draw, fill={rgb:blue,1;white,3}] (v)
	{\large $v$};
	\node[above=0.01cm of v, align=center] (vL) 
	{Latent\\ Shape};
	\path[every node/.style={anchor=south}]
	(inshape) edge [->,>=stealth'] node [midway,above,align=center] 
	{\large $f_{v}$} (v);
	\node[right=2.2cm of v, yshift=0.0cm, xshift=0.0cm] (M) 
	{ \includegraphics[width=.14025\textwidth]{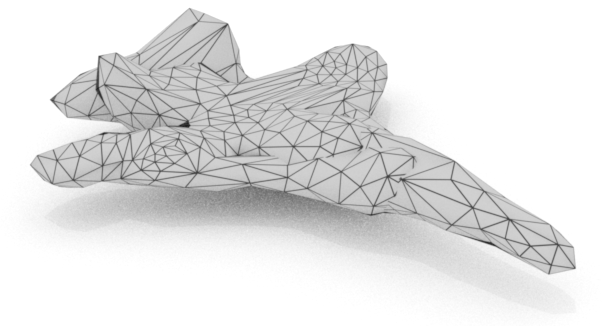} };

	\definecolor{littleblue}{RGB}{137,213,252};
	
	\node[right=0.9cm of v, yshift=0.0cm, xshift=0.0cm, rectangle, rounded corners, draw,fill=littleblue] (delta) 
	{ $\delta$ };
	\node[main node, below=0.1cm of delta] (template)
	{ \includegraphics[width=.04\textwidth]{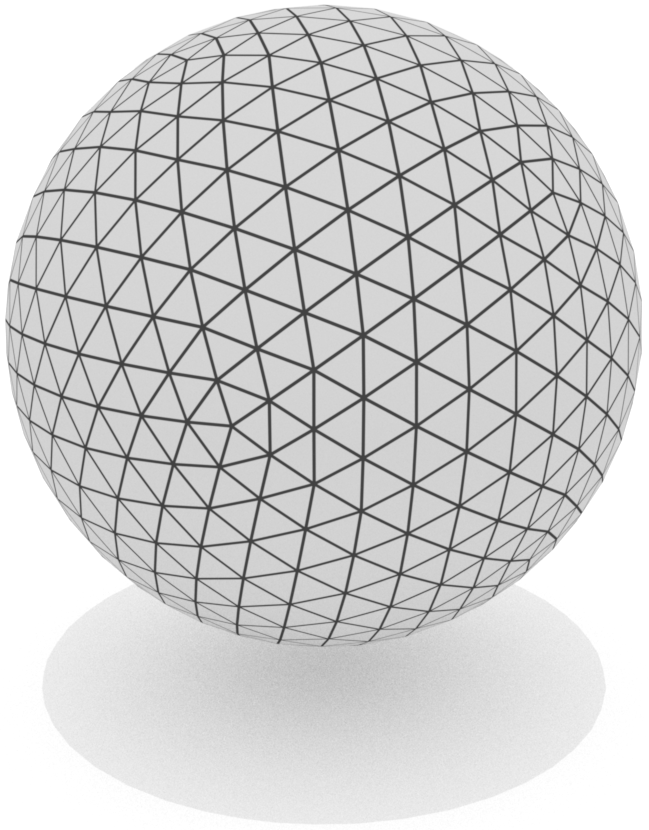} };
	\node[main node, left=0.0cm of template,xshift=2mm](templatelabel)
	{ $S_T$ };
	
	\node (betweendeltaMtilde) at ($(delta.east)!0.5!(M.west)$) {};
	\path let \p1 = (betweendeltaMtilde.center), \p2 = (template.center) 
	in node (templatecorner) at (\x1,\y2) { };
	\draw[-latex,black,-] (template.east) -- (templatecorner.center);
	\draw[-latex,black,-] (templatecorner.center) -- (betweendeltaMtilde.center);
	
	\path[every node/.style={anchor=south}]
	(delta) edge [->,>=stealth'] node [pos=0.5,above,align=center] 
	{\large $+$} (M);	
	
	\path[every node/.style={anchor=south}]
	(v) edge [->,>=stealth'] node [midway,above,align=center] 
	{\large $f_{\delta}$} (delta);	
	\node[main node,below=0.15cm of inshape] (inxit) %
	{ {\transparent{0.7}\includegraphics[width=.105\textwidth]{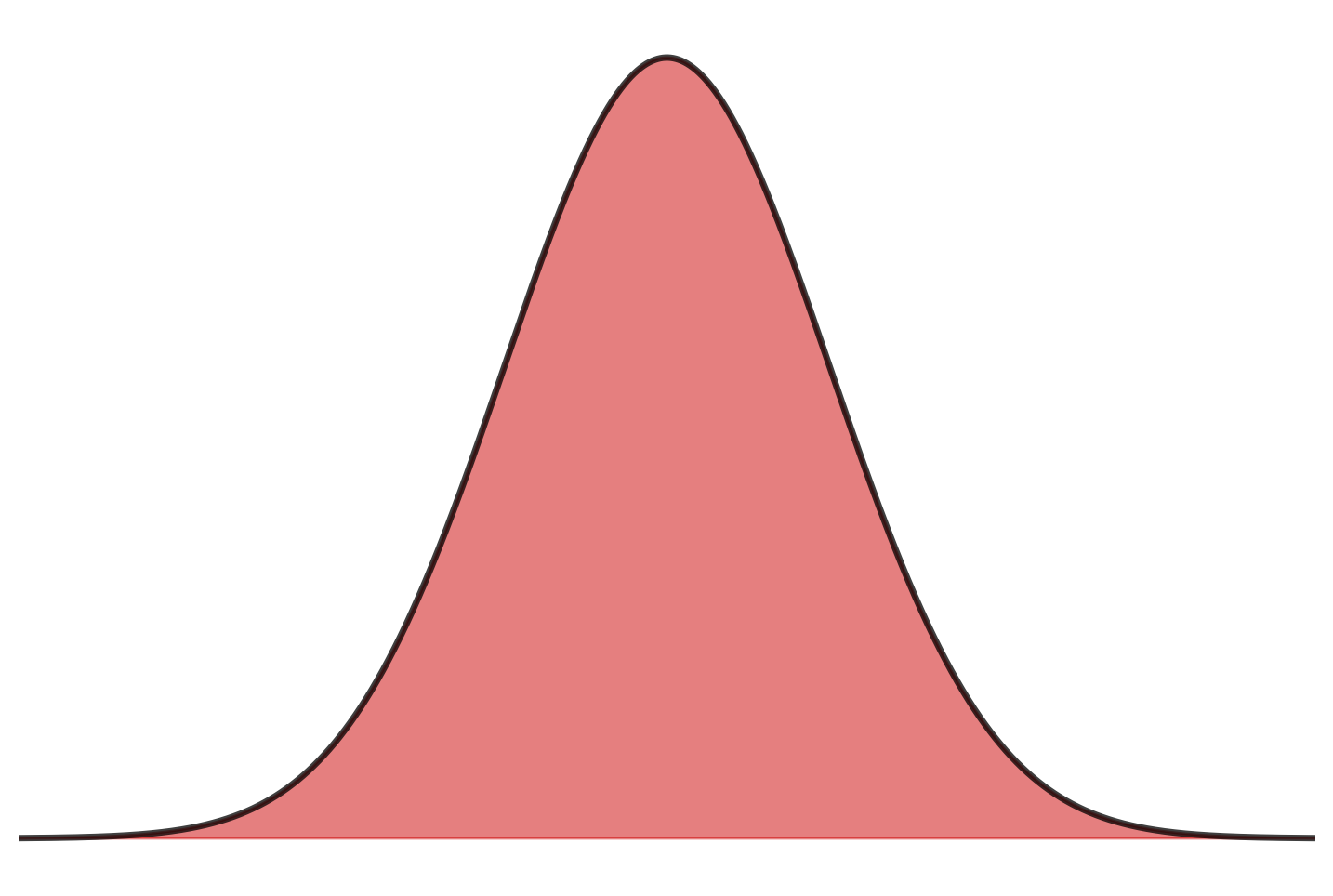} } };
	\node[yshift=0.7cm] at (inxit) (inxitLL) {Latent Texture};
	\node (inxitL) at (inxit) {\large $ \xi_T\sim \mathcal{N}(0,I) $};
	\node[main node,above=0.4cm of inshape] (inxip) %
	{ {\transparent{0.7}\includegraphics[width=.105\textwidth]{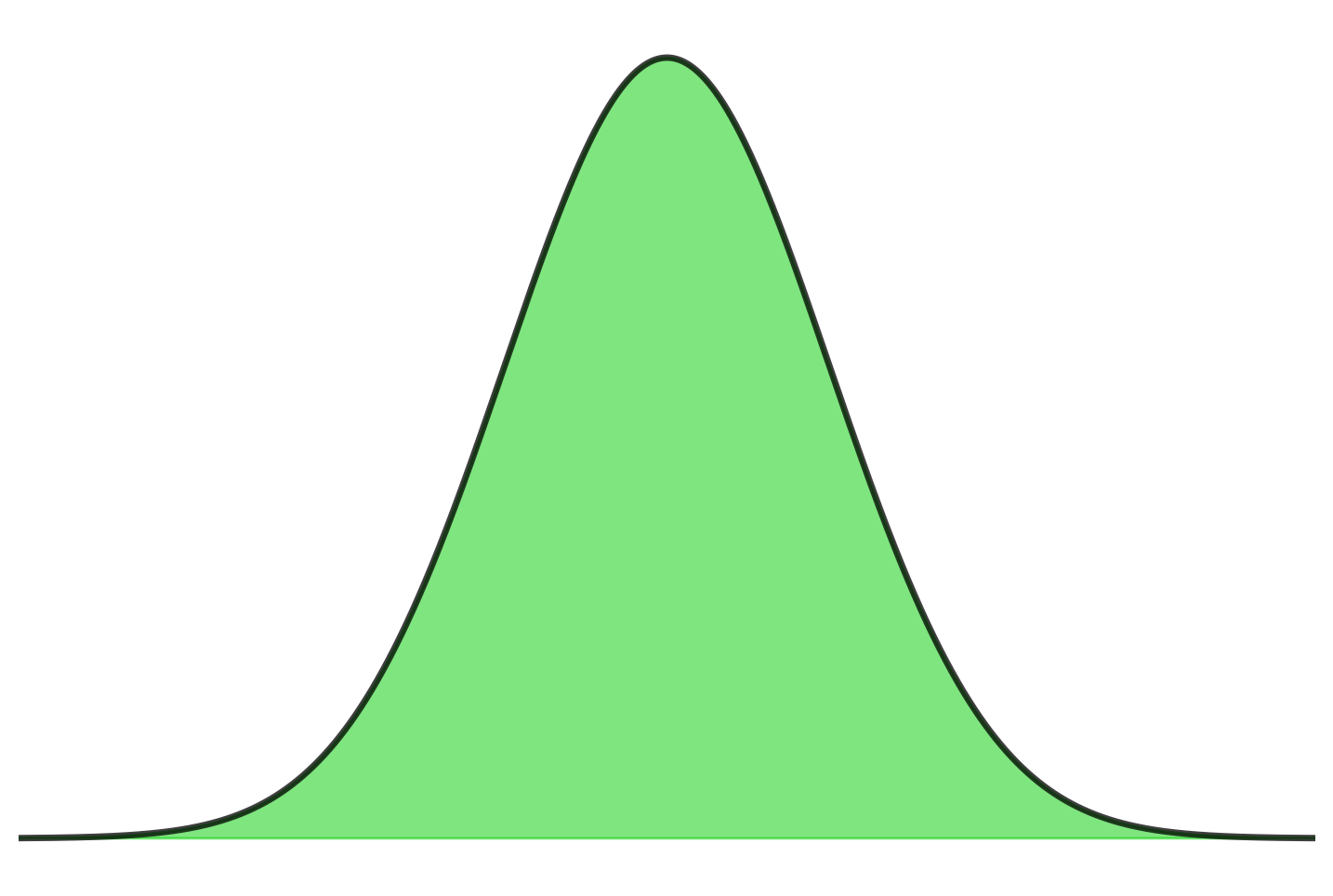} } };
	\node (inxipL) at (inxip)
	{\large $\, \xi_p\sim \mathcal{N}(0,I) $};
	\node[yshift=0.7cm] at (inxip) (inxipLL) 
	{Latent Pose};
	\node[main node, right=2.1cm of inxipL, rectangle, rounded corners, draw, fill={rgb:green,0.5;white,2}] (ET)
	{\large $E$};
	\node[main node, below=0.01cm of ET, rectangle, rounded corners,yshift=1mm] (frust)
	{\includegraphics[width=.04\textwidth]{temp/frust_1.pdf}};
	\node[above=0.01cm of ET,yshift=-0.115cm] (ETL) 
	{Rigid Pose};
	\path[every node/.style={anchor=south}]
	(inxipL) edge [->,>=stealth'] node [midway,above,align=center] {\large $f_{p}$} (ET);
	\node[above=0.001cm of M, align=center, yshift=-0.315cm] (ML) 
		{Canonical Shape $M$};
	\node[right=6.7cm of ET, yshift=-0.0cm,xshift=0.0cm] (ME) 
		{ \includegraphics[width=.125\textwidth]{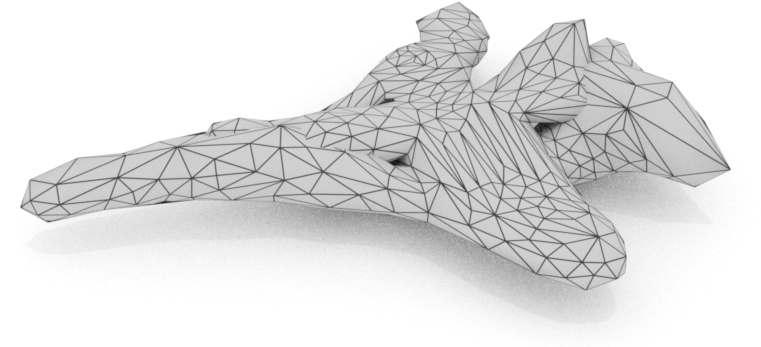} };
	\node[above=0.001cm of ME, align=center, yshift=-0.29cm] (MEL) 
		{Scene-Space\\ Shape $M_E$};	
	\path[every node/.style={anchor=south}]
	(ET) edge [->,>=stealth'] node [midway,above,align=center,pos=0.8] 
	{\large $f_E$} (ME);
	\path let \p1 = (M), \p2 = (ET) 
	in node (lolytho2) at (\x1,\y2) { };
	\draw[-latex,black,-] (ML.north) -- (lolytho2.center);
	
	\node[right=4.785cm of M, yshift=-0.0cm,xshift=0.0cm] (render) 
	{ \includegraphics[width=.1\textwidth]{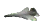} };
	\node[yshift=0.95cm,align=center] at (render) (renderL) 
	{Generated\\ Render $\widetilde{I}$ };
	\path let \p1 = (render), \p2 = (ME) 
	in node (fakefake2) at (\x2,\y1) {};
	\path[every node/.style={anchor=south}]
	(fakefake2.center) edge [->,>=stealth'] node [midway,above,align=center] {\large $\mathcal{R}$} (render);
	
	\path let \p1 = (ME), \p2 = (inxit) 
		in node (newtp) at (\x1,\y2) {};
	\node[main node,xshift=-0.45cm] at (newtp) (Thelp) 
	{};
	
	\node[main node] at (newtp) (T) %
	{};
	\path let \p1 = (M.center), \p2 = (T) 
	in node (lolythomt) at (\x1,\y2) { };
	\draw[-latex,black,-] (M.south) -- (lolythomt.center);
	\node[align=center,yshift=0.5cm] at (T) (TL) {Texture $T$};

	\path let \p1 = (ME), \p2 = (TL.north) 
	in node (fakefake) at (\x1,\y2) {};
	\draw[-latex,black,-] (ME.south) -- (fakefake.center);
	
	\def\texwidth{0.7}
	
	\node (rect1) at (T) [rectangle,draw,thin,minimum width=\texwidth cm,minimum height=0.08cm,fill=gray,outer sep=0pt, inner sep=0pt,yshift=0.2cm] {};
	\node (rect2) [rectangle,draw,thin,minimum width=\texwidth cm,minimum height=0.08cm,fill=black,outer sep=0pt, inner sep=0pt,below=0.05cm of rect1] {};
	\node (rect3)[rectangle,draw,thin,minimum width=\texwidth cm,minimum height=0.08cm, fill={rgb:black,1;white,3},outer sep=0pt, inner sep=0pt,below=0.05cm of rect2] {};
	\node (rect4)  [rectangle,draw,thin,minimum width=\texwidth cm,minimum height=0.08cm, fill={rgb:black,1.5;white,3},outer sep=0pt, inner sep=0pt,below=0.05cm of rect3] {};
	\node (rect5)  [rectangle,draw,thin,minimum width=\texwidth cm,minimum height=0.08cm,fill={rgb:black,4;white,3},outer sep=0pt, inner sep=0pt,below=0.05cm of rect4] {};
	\node (rect6)  [rectangle,draw,thin,minimum width=\texwidth cm,minimum height=0.08cm,fill={rgb:black,0.5;white,1},outer sep=0pt, inner sep=0pt,below=0.05cm of rect5] {};
	
	\node[main node,right=6.97cm of inxit] (TI) %
	{\includegraphics[width=.05\textwidth]{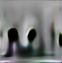}};
	\node[main node, above=1mm of TI,yshift=-0.3cm] (TIlabel)
	{ Texture Image $T_I$ };

	\path[every node/.style={anchor=south}]
	(inxitL) edge [->,>=stealth'] node [midway,below,align=center,pos=0.92] {\large $f_{T}$} (TI);
	\path[every node/.style={anchor=south}]
	(TI) edge [->,>=stealth'] node [midway,below,align=center,pos=0.5] {Sample} (Thelp);
	
	\draw [decorate,decoration={brace,amplitude=5pt},xshift=50mm,yshift=0pt] 
	($(rect1.north east)+(0.07, 0.04)$) -- ($(rect6.south east)+(0.07, -0.04)$) 
	node [black,midway,align=left,xshift=5.5mm] { \footnotesize Per\\[-2mm] \footnotesize Node };
	\end{tikzpicture}%
	\caption{The architecture of our \textit{shape-to-image} translation function, $\fsi$. Starting from an input shape $P$ and sampled latent vectors, $\xi_T$ and $\xi_p$, we first encode $P$ into its latent form $v$, which is used to obtain the canonical shape, $M$, via adding the nodal offset $\delta$ to the template $S_T$.
	Conditioned on $M$, the latent texture $\xi_T$ is decoded into the UV texture image $T_I$, which is sampled to obtain the texture $T$ as nodal colours. 
	 The rigid pose is independently decoded into a Euclidean transform, $E$, and applied to $M$, producing $M_E$. Finally, $M_E$ and $T$ are differentiably rendered into the image, $\widetilde{I}$. } %
	\label{fig:fsi}
	\end{figure*}%
	\begin{figure*}
		\centering
	\begin{tikzpicture}[main node/.style={},square/.style={regular polygon,regular polygon sides=4}]
	\node[main node,square,inner sep=-0.35em,draw,xslant=0.0, yslant=0.0, anchor=south] (inimg)
	{\includegraphics[width=0.05\textwidth]{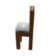}};
	\node[yshift=0.75cm] at (inimg) (inimgL) {Input Image};
	\node[yshift=-0.8cm] at (inimg) (inimgLB) { \large $I$};
	\node[main node,right=0.05cm of inimg] (cnn) %
	{ {\transparent{1.0}\includegraphics[width=.09\textwidth]{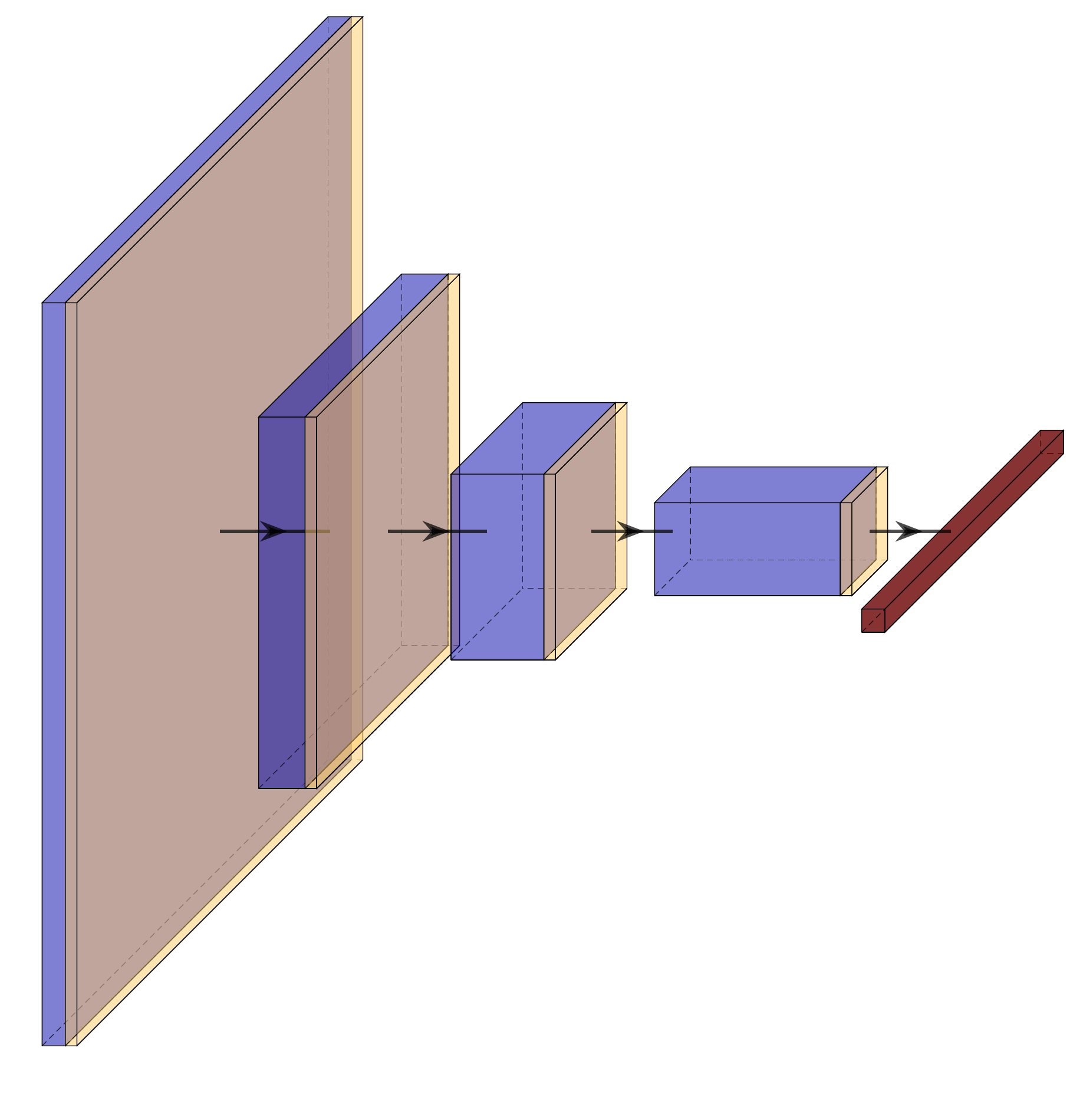} } };
	\node[yshift=0.85cm] at (cnn) (cnnlab) { \large $h_q$ };
	\node[yshift=-0.4cm,xshift=0.6cm] at (cnn) (q) { \large $q$};
	\node[xshift=1.0cm] at (cnn) (fcnn) 
	{};
	\node[xshift=0.4cm] at (fcnn) (centerpiece) 
	{};
	\node[yshift=1.85cm] at (centerpiece) (centerpieceup) 
	{};
	\node[yshift=-2.07cm] at (centerpiece) (centerpiecedown) 
	{};
	\draw[-latex,black,-] (fcnn.center) -- (centerpiece.center);
	\draw[-latex,black,-] (centerpiece.center) -- (centerpieceup.center);
	\draw[-latex,black,-] (centerpiece.center) -- (centerpiecedown.center);
	\node[main node, right=1.295cm of centerpiece, circle, draw, fill={rgb:blue,1;white,3}] (v)
	{\large $\widetilde{v}$};

	\path[every node/.style={anchor=south}]
	(centerpiece.center) edge [->,>=stealth'] node [midway,above,align=center] {\large $h_v$} (v);
	\node[main node, right=1.35cm of centerpiecedown, rectangle, rounded corners, draw, fill={rgb:green,0.5;white,2}] (ET)
		{\large $\widetilde{E}$}; %
		\node[main node, below=0.01cm of ET, rectangle, rounded corners,yshift=1mm] (frust)
		{\includegraphics[width=.04\textwidth]{temp/frust_1.pdf}};
	\node[above=0.1cm of ET,yshift=-0.115cm,align=center] (ETL) 
	{Rigid\\ Pose};
	\path[every node/.style={anchor=south}]
	(centerpiecedown.center) edge [->,>=stealth'] node [midway,above,align=center] {\large $h_p$} (ET);
	\node[above=0.01cm of v, align=center] (vL) 
	{Latent\\ Shape};
	\path let \p1 = (vL), \p2 = (centerpieceup) 
	in node (lolytho9) at (\x1,\y2) { };
	\draw[-latex,black,-] (vL.north) -- (lolytho9.center);
	\definecolor{littleblue}{RGB}{137,213,252};
	
	\node[right=1.0cm of v, yshift=0.0cm, xshift=0.0cm, rectangle, rounded corners, draw,fill=littleblue] (delta) 
		{ $\widetilde{\delta}$ };
	\node[right=2.5cm of v, yshift=0.0cm, xshift=0.0cm] (M) 
		{ \includegraphics[width=.051175\textwidth]{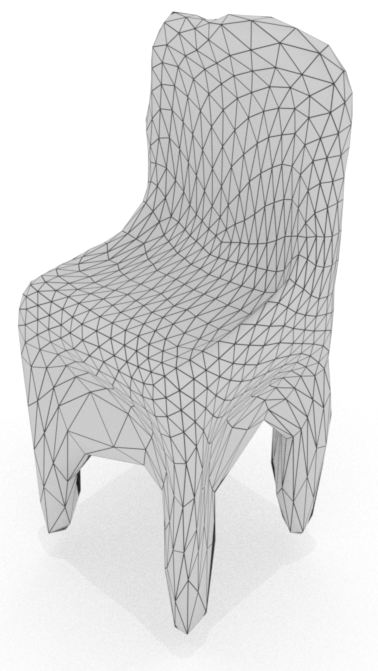} };
			
	\node[main node, below=0.1cm of delta] (template)
		{ \includegraphics[width=.04\textwidth]{temp/sphere.png} };
	\node[main node, left=0.0cm of template,xshift=2mm](templatelabel)
		{ $S_T$ };
		
	\node (betweendeltaMtilde) at ($(delta.east)!0.5!(M.west)$) {};
	\path let \p1 = (betweendeltaMtilde.center), \p2 = (template.center) 
	in node (templatecorner) at (\x1,\y2) { };
	\draw[-latex,black,-] (template.east) -- (templatecorner.center);
	\draw[-latex,black,-] (templatecorner.center) -- (betweendeltaMtilde.center);
		
	\path[every node/.style={anchor=south}]
	(delta) edge [->,>=stealth'] node [pos=0.5,above,align=center] 
	{\large $+$} (M);	
	
	\path[every node/.style={anchor=south}]
	(v) edge [->,>=stealth'] node [midway,above,align=center] 
	{\large $f_{\delta}$} (delta);	
	\node[above=0.001cm of M, align=center, yshift=-0.27315cm] (ML) 
	{Canonical Shape $\widetilde{M}$};
	\path let \p1 = (M), \p2 = (ET) 
	in node (lolytho) at (\x1,\y2) { };
	\draw[-latex,black,-] (M.south) -- (lolytho.center);
	\draw[-latex,black,-] (ET.east) -- (lolytho.center);
	\node[right=1.5cm of lolytho, yshift=-0.0cm, xshift=-0.6cm] (ME) 
	{  };
	\node[above=0.05cm of ME, yshift=-0.9cm,xshift=0.5cm](MEimg){
		\includegraphics[width=.0572\textwidth]{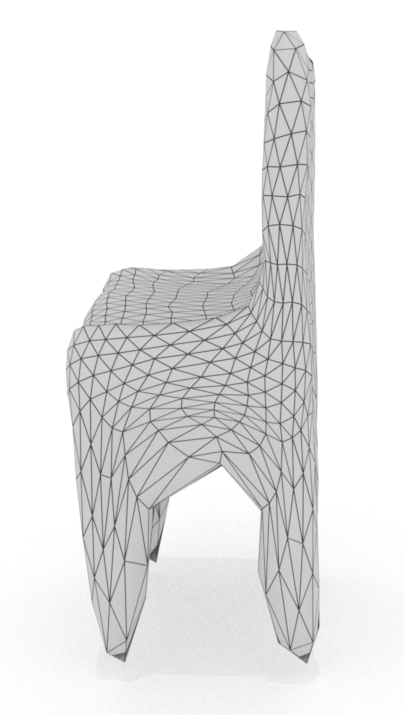} };
	\node[xshift=0.29cm] (MEimg2) at (MEimg.center){};
	\node[above=0.001cm of MEimg, align=center, yshift=-0.29cm] (MEL) 
	{Scene-Space\\[0.041in] Shape $\widetilde{M}_E$};
	\path[every node/.style={anchor=south}]
	(lolytho.center) edge [->,>=stealth'] node [midway,above,align=center] 
	{\large $f_{E}$} (ME);	
	\node[below=1.7cm of inimgLB, align=center, yshift=0.0cm] (belowI) 
	{};
	\path let \p1 = (belowI), \p2 = (ME) 
	in node (Uint) at (\x2,\y1) {};	
	\node[right=0.285cm of Uint, align=center, yshift=0.0cm, circle] (U) 
	{};
	\node[above=0.1cm of Uint, align=center, xshift=-1.75cm, yshift=-0.27cm] (unproj) 
	{};
	\draw[-latex,black,-] (inimgLB.south) -- (belowI.center);
	\draw[-latex,black,-] (belowI.center) -- (Uint.center);
	\path[every node/.style={anchor=south}]
	(Uint.center) edge [>=stealth'] node [midway,above,align=center] 
	{} (U.center);
	\node[right=0.9cm of U, align=center, yshift=0.0cm] (Uright) 
	{};
	\draw[-latex,black,-] (U.center) -- (Uright.center);
	\path let \p1 = (centerpieceup), \p2 = (Uright) 
	in node (ta) at (\x2,\y1) {};
	
	\draw[-latex,black,-] (centerpieceup.center) -- (ta.center);	
	\definecolor{loragne}{RGB}{255,219,135};
	\path let \p1 = (Uright), \p2 = (MEimg) 
	in node[fill=loragne,circle,draw] (ta2) at (\x1,\y2) {$U$};
	\draw[-latex,black,-] (ta2.north) -- (ta.center);
	\path[every node/.style={anchor=south}]
	(Uright.center) edge [->,>=stealth'] node [midway,right,align=center] 
	{Unproject} (ta2);	
	\path[every node/.style={anchor=south}]
	(MEimg2) edge [->,>=stealth'] node [midway,right,align=center] 
	{} (ta2);	

	\node[right=1.1cm of ta, align=center, xshift=-0.0cm, yshift=0.0cm, rectangle, rounded corners, fill={rgb:red,1;white,2}, draw] (xiT) 
	{\large $\widetilde{\xi}_T$};

	\node[main node, below=0.25cm of xiT, rectangle, rounded corners, draw, fill={rgb:green,0.5;white,2},text opacity=1,fill opacity=0.5,xshift=4mm,yshift=-10mm] (ET2)
	{\large $\widetilde{E}$};

	\node[right=0.0cm of xiT, align=center, xshift=-0.0cm, yshift=0.0cm] (xiTL) 
	{Latent\\ Texture};	
	\path[every node/.style={anchor=south}]
	(ta.center) edge [->,>=stealth'] node [midway,below,align=center] 
	{\large $h_T$} (xiT);

	\node[main node, right=2.7mm of ET2, circle, draw, fill={rgb:blue,1;white,3},xshift=0mm,yshift=-0mm] (v2)
		{\large $\widetilde{v}$};

	\node[right=2.7mm of v2, align=center, xshift=-0.0cm, yshift=0.0cm, rectangle, rounded corners, fill={rgb:red,1;white,2}, draw] (xiT2) 
		{\large $\widetilde{\xi}_T$};

	\node[minimum height=0.37in, minimum width=1.1in, rounded corners, rectangle, thick, black, xshift=0.5mm, draw] at (v2) (box) {};

	\node [below of=box,yshift=0.15cm,align=center]{Inferred Graphics Code}; 

	\end{tikzpicture}
	\caption{The architecture of the \textit{image-to-shape} translation function $\fis$. Given an input image $I$, we process it into an encoding $q$, which is subsequently used to estimate latent shape $\widetilde{v}$ and rigid pose $\widetilde{E}$. 
	Then, $\widetilde{v}$ is decoded into a nodal perturbation, $\widetilde{\delta}$, and applied to the template, $S_T$, to get $\widetilde{M}$, and $\widetilde{E}$ is used to transform $\widetilde{M}$ into $\widetilde{M}_E$. 
	Lastly, the pixel values of $I$ are unprojected onto $\widetilde{M}_E$ to obtain nodal colours $U$, which are mapped to $\widetilde{\xi}_T$. %
	The final output is the inferred graphics code 
	$\widetilde{\mathfrak{C}} = (\widetilde{E}, \widetilde{v}, \widetilde{\xi}_T)$,
	which can be decoded by %
	$\fsi$.
	}
	\label{fig:fis}
\end{figure*} 

\section{Model}
\label{sec:model}

\subsection{Overview and Definitions} 
\label{sec:model:overview}

Our model consists of two main algorithms, 
which are composed to compute our cycle-consistent training objective,
as shown in Fig.\ \ref{fig:cycles}:
the shape-to-image and image-to-shape functions, 
$f_{S \rightarrow I}$
and
$f_{I \rightarrow S}$
(depicted in Fig.\ \ref{fig:fsi} and \ref{fig:fis}, respectively).
Given a novel shape, 
	$f_{S \rightarrow I}$ samples a texture and viewpoint for a novel shape, then renders an image,
while $f_{I \rightarrow S}$ performs 3D shape, texture, and viewpoint inference from an input image (inverse graphics).
We define latent vectors 
$\xi_T\sim\mathcal{N}(0,I)$, 
$\xi_p\sim\mathcal{N}(0,I)$, 
and $v$,
for the texture (3D appearance), 
Euclidean pose, and 
non-rigid shape, respectively.	
Our shape representation utilizes a template mesh, 
$S_T = (\mathcal{V},\mathcal{E},\mathcal{F})$, 
written as a tuple of vertices, edges, and faces, 
which we deform to match either a given input shape or image, 
through a latent shape vector $v$.
The texture $T\in\mathbb{R}^{|\mathcal{V}|\times 3}$ is simply the colour values per vertex, 
on the template $S_T$.
The rigid pose transform (mapping object to scene space), $E=(R,t)$, 
consists of a global rotation $R\in SO(3)$ and translation $t\in\mathbb{R}^3$.
The rotation matrix $R$ is computed via the 
Tait-Bryan angle \cite{diebel2006representing} vector representation $r\in\mathbb{R}^3$, and then
 converted to $R$.

Input shapes are assumed to be point clouds $P\in\mathbb{R}^{N_S \times 6}$ 
in approximate rigid alignment with $N_S$ points,
along with their associated surface normals 
(which we convert to a canonical template representation $M$), 
while 2D input images are denoted by $I$. 
We use $\widetilde{v}, \widetilde{M}, \widetilde{\xi}_T, \widetilde{E}$, and $ \widetilde{I}$ to denote the 
shapes (latent and mesh), textures, poses, and renders from a single modality translation application,
respectively,
and $\widehat{v}, \widehat{M}, \widehat{\xi}_T, \widehat{E}$, and $\widehat{I}$  for those from a second translation (i.e., reconstructions).

The core of our approach is to enforce that
(a)
$(\widetilde{M}, \widetilde{\xi}_T, \widetilde{E}) = f_{I \rightarrow S}(I) $
and
$\widetilde{I} = f_{S \rightarrow I}(M, \xi_T, E) $
produce ``realistic'' outputs, via adversarial distribution matching,
and
(b)
cycle-consistency is upheld, meaning
$ I \approx \widehat{I} = f_{S \rightarrow I}(f_{I \rightarrow S}(I)) $
and
$(P, \xi_T, E) \approx 
(\widehat{M}, \widehat{\xi}_T, \widehat{E}) = 
f_{I \rightarrow S}(f_{S \rightarrow I}(M, \xi_T, E))  $,
where $\xi_T$ and $E$ are sampled from their respective priors.
We call these two cycles the \textit{vision cycle} (VC) and \textit{graphics cycle} (GC), respectively.
For further details, we refer the reader to the appendix (Sec.\ \ref{suppsec:model}).

\subsection{Shape-to-Image Translation}
\label{sec:model:shape2img}

Consider an oriented point set, $P$, which we intend to render into an image (see Fig.\ \ref{fig:fsi}). 
From Gaussian priors, we start by sampling a latent texture, $\xi_T$, and pose, $\xi_p$, 
and decode the latter into a Euclidean transform $E = (R,t) = f_p(\xi_p)$. 
We first map the input, $P$, into a canonical parametrization, by deforming the template mesh, $S_T$. This helps ensure that our vertex-wise texture is always computed in the same space, and that there is an approximate correspondence between nodes across input shapes.
We thus infer the latent shape vector via $v = f_v(P)$, 
where $f_v$ is implemented as a standard PointNet \cite{qi2017pointnet}, 
able to handle a dynamic number of points.
Together, the tuple $\mathfrak{C} = (E, v, \xi_T)$ may be considered the ``graphics code'' of our model,
encompassing shape, appearance, and pose, all in 3D.
With some abuse of notation, this code $\mathfrak{C}$ can be viewed as the input to $\fsi$ and the output of $\fis$.\footnote {We use the latent vector $v$ and the corresponding mesh $M$ interchangeably to refer to a 3D shape.}

Then, we decode the latent shape $v$ into a set of nodal offsets $\delta = f_\delta(v) \in\mathbb{R}^{|\mathcal{V}|\times 3}$,
so that the vertices of the deformed shape may be written as
$M = \mathcal{V} + \delta$ (e.g., as in \cite{kanazawa2018learning}). 
This canonically parametrized shape, $M$, is in canonical (object-centred) coordinates, 
as well.
As in an extrinsic camera transform,
we apply the Euclidean transform $E=(R,t)$ to obtain the scene-space  shape
via
$M_E = f_E(M,E) = MR + t$, where $R$ and $t$ are applied to every vertex.

Next, we decode the texture, conditionally on the shape: $ T = f_T(\xi_T, M) $.
Even for a fixed object category, there may be sufficient non-rigid deformation 
that a given texture will not be applicable to every shape; 
conditioning on the canonical shape, $M$, 
	allows the networks to adjust to such variations.
We note that the texture decoder, $f_T$, proceeds by first mapping $\xi_T$ to a texture image $T_I$ (with an upsampling convolutional network), and then uses the fixed UV texture coordinates of the spherical template to obtain nodal pixel values by bilinear sampling \cite{jaderberg2015spatial} from $T_I$. 
Finally, we apply the SoftRas differentiable renderer \cite{liu2019soft} 
to generate an image from the textured shape,
via $\widetilde{I} = \mathcal{R}(M_E, T)$.
We use constant ambient lighting and a camera with fixed intrinsics throughout the paper.
This complete mapping constitutes the shape-to-image function 
$\widetilde{I} = f_{ S \rightarrow I}(f_v(P), \xi_T, E)$, as depicted in Fig.\ \ref{fig:fsi}.

\subsection{Image-to-Shape Translation} %
\label{sec:model:img2shape}

Now consider an input image $I$, 
from which we wish to extract a textured 3D shape representation (see Fig.\ \ref{fig:fis}).
We first pass the image through a ResNet-18 \cite{he2016deep} 
to extract a vector image encoding, $q = h_q(I)$.
The pose and latent shape are then estimated from $q$, as
$\widetilde{E} = h_p(q)$ and $\widetilde{v} = h_v(q)$.
As in shape-to-image translation, we deform the template, 
as
$\widetilde{M} =  \mathcal{V} + f_\delta(\widetilde{v})$,
and place it in scene coordinates:
$ \widetilde{M}_E = f_E(\widetilde{M} ,\widetilde{E}) $.

We next estimate the full 3D texture from the image.
Since our inference algorithm is designed to ``explain'' the observed image as a reprojection
of the predicted 3D shape,
we can directly detect the colour each vertex \textit{should} be,
based on which pixel it is projected to in the original image.
Similar to recent work \cite{henderson2020leveraging,kanazawa2018learning,tulsiani2020implicit},
we use this colour information to help infer the latent texture;
but, note that these values cannot be used directly, %
since the generative shape-to-image function $\fsi$ expects $\xi_T$ to be Gaussian.
Let $U = \mathcal{P}_U(I,\widetilde{M}_E) \in \mathbb{R}^{|\mathcal{V}|\times 3}$
be the \textit{pixel unprojection} of the image $I$ onto the transformed shape, $\widetilde{M}_E$.
The $i^\text{th}$ subvector of $U$ is the colour 
$I(\widetilde{x}_i)$, where $\widetilde{x}_i$ is the pixel projection of the 
$i^\text{th}$ template vertex (from $\widetilde{M}_E$).
The inferred latent texture encoding is finally computed 
via %
$\widetilde{\xi}_T = h_T(q,\widetilde{v},U)$,
which can be decoded into a complete texture  
$\widetilde{T} = f_T(\widetilde{\xi}_T, \widetilde{M})$. %

Following prior work involving 3D rigid pose inference 
\cite{Tulsiani_2018_CVPR,kulkarni2019canonical},
we also utilize a set of $n_h$ \textit{pose hypotheses} when performing inference, 
to avoid getting trapped in local minima.
Thus, any call to $f_{I\rightarrow S}$ produces only a single canonical shape $\widetilde{M}$, but $n_h$ rotations $\widetilde{R}_i$, translations $\widetilde{t}_i$, scene-space shapes $\widetilde{M}_{E,i}$, textures $\widetilde{\xi}_{T,i}$ (due to the pose dependence of the unprojection), and pose probabilities $p_i$ for $i\in[1,n_h]$.  
In either cycle, 
these outputs are processed independently until their use in the objective function,
similar to \cite{Tulsiani_2018_CVPR,kulkarni2019canonical}.

\subsection{Cycle-Consistent Loss Objective}
\label{sec:model:lossobj}

Similar to cycle-consistent GANs, our overall loss function consists of adversarial distribution matching and reconstruction consistency losses from the two cycles formed by $\fis$ and $\fsi$ (depicted in Fig.\ \ref{fig:cycles}):
\begin{equation}
	\mathcal{L} = \mathcal{L}_G + \mathcal{D}_G + \mathcal{L}_V + \mathcal{D}_V +  \mathcal{L}_R,
\end{equation}
where the reconstruction loss, $\mathcal{L}_G $,  and distribution matching loss, $\mathcal{D}_G$, form the \textit{graphics cycle} objective
(from applying $\fsi$, then $\fis$), 
while
$\mathcal{L}_V$ and $\mathcal{D}_V$ do so for the \textit{vision cycle} 
(applying $\fis$, then $\fsi$).
The objective term $\mathcal{L}_R$ is comprised of regularization losses.
While we expand on each term below,
due to space constraints, 
we refer the reader to the appendix for additional details (Sec.\ \ref{suppsec:losses}). 

\subsubsection{Graphics Cycle Losses}

\paragraph{Distribution Matching Loss $ \boldsymbol{\mathcal{D}}_{\boldsymbol{G}} $.}%
\label{sec:gcldm}
We impose distribution matching losses on the generated render $\widetilde{I}$,  texture image $T_I$ (which is sampled to obtain $T$), and the Euclidean poses, $E$. 
The losses on both $\widetilde{I}$ and $T_I$ are implemented 
with convolutional WGAN-GP image critics \cite{arjovsky2017wasserstein, gulrajani2017improved}, $C_I$ and $C_T$ respectively.
As in \cite{chen2019learning}, 
	we train $C_T$ using inferred textures $\widetilde{T}$ from the vision cycle (VC).
Similarly, we use the pose distribution from the VC to impose a loss on $E$.
While one might expect to learn the correct rigid pose distribution from the image adversary, 
	we found this was more stable. %
Since the Euclidean transform elements, $R$ and $t$, are low dimensional, 
we use a Sinkhorn distance \cite{cuturi2013sinkhorn} 
(via GeomLoss \cite{feydy2018interpolating}) 
to match their distributions. 
We maintain a buffer of recently inferred $\widetilde{E}$ values from the VC, matching each batch from the graphics cycle (GC) to it. 
We use the $L_2$ metric for $t$ and the geodesic distance on $SO(3)$, written $d_{R}$, 
for $R$ \cite{huynh2009metrics}.

\paragraph{Cyclic Consistency Loss $ \boldsymbol{\mathcal{L}}_{\boldsymbol{G}} $.}%
After computing $\widetilde{I} = f_{S \rightarrow I}(f_v(P), \xi_T, E)$, we obtain $\widehat{v}$, $\widehat{M}$, $\widehat{E}$, and $\widehat{\xi}_T$ from $\fis(\widetilde{I} )$.
Consistency in shape, texture, and pose is enforced by matching $E$ and $\widehat{E}$, $\xi_T$ and $\widehat{\xi}_T$, and $(v, M)$ and $(\widehat{v}, \widehat{M})$, respectively, all as vectors 
(except for $R$ and $\widehat{R}$, which use $d_R$).
We also use a metric between the input %
$P$ and the canonically deformed template shape $M$, 
consisting of a Chamfer and a normals-matching loss, 
as in \cite{gkioxari2019mesh}.

\subsubsection{Vision Cycle Losses}
\paragraph{Distribution Matching Loss $ \boldsymbol{\mathcal{D}}_{\boldsymbol{V}} $.}%
There are two distribution-matching losses in the VC.
The first is on the inferred shape $(\widetilde{v}, \widetilde{M})$, which we implement as a vector critic $C_s$. 
This loss encourages the canonical shape $\widetilde{M}$, inferred from the image $I$, to look like a deformed template, obtained by encoding a \textit{real} 3D shape, $P$, in the GC.
The second is on the inferred latent texture, $\xi_T$, which we enforce to be Gaussian, 
consistent with our sampling in the GC.
Since $\xi_T$ is relatively low-dimensional, we follow recent work in generative models and apply the sliced Wasserstein distance (SWD) \cite{deshpande2018generative,rabin2011wasserstein} to push $\xi_T$ to be normally distributed, due to its stability and dearth of (hyper-)parameters.
As noted in Sec.\ \ref{sec:gcldm}, we do not use a critic on $\widetilde{E}$, instead relying on the rigid poses inferred in the VC to be realistic. %

We remark here that all adversarial critics (i.e.,  those applied on the 
inferred shapes, 
$C_s(\widetilde{v},\widetilde{M})$, 
 generated textures, 
 $C_T(\widetilde{T}_I)$, 
 and sampled renders, 
 $C_I(\widetilde{I})$) 
use the WGAN-GP loss formulation \cite{gulrajani2017improved}, with
an additional ``drift'' penalty \cite{karras2018progressive},
to prevent losses from growing too large and overwhelming other generator terms.

\paragraph{Cyclic Consistency Loss $ \boldsymbol{\mathcal{L}}_{\boldsymbol{V}} $.}%
The only reconstruction loss in the VC is the image-to-image distance 
	between $I$ and the reconstructed render $ \widehat{I} =  \fsi(\fis(I)) $. 
We use a combination of an $L_1$ pixelwise loss and a perceptual metric, computed using the intermediate feature maps of the image critic $C_I$ 
(as in, e.g., \cite{larsen2016autoencoding,wang2018high}).
The loss is weighted as
an expectation 
over the pose hypothesis probabilities, $p_i$. %

\subsubsection{Regularization Losses}

As in most learning models that operate on meshes, 
	we apply geometric regularizations to prevent pathological shape outputs 
	(e.g., rampant self-intersection \cite{henderson2020leveraging} and
	         ``flying'' vertices \cite{wang2018pixel2mesh}). %
In particular, we follow prior work \cite{kato2018neural, wang2018pixel2mesh,liu2019soft,gkioxari2019mesh} and apply a 
Laplacian, %
flatness,  %
and 
edge length losses %
to the canonical meshes $M$ and $\widetilde{M}$. 
We add an additional term for $\delta$ and $\widetilde{\delta}$, 
	penalizing the per-face variance of the nodal perturbations.

We also regularize the exploration of the Euclidean pose hypotheses.
As in \cite{kulkarni2019canonical}, 
	we encourage diversity in the rotations by maximizing the expected pairwise distance between the rotational hypotheses, via
	$ L_{RD} = \sum_{i,j} p_i p_j d_R(R_i, R_j) $.
For $t$ and $\xi_T$ in the VC, we penalize the distance of each hypothesis 
to the ``best'' (highest probability) vector, instead. 
This helps decrease ``drift'' in the inferred translation $\widetilde{t}$, which reduces instability 
(e.g., due to the mesh leaving the view frustum).
For the latent texture $\widetilde{\xi}_T$, we encourage consistent textures regardless of the estimated pose.

Finally, while texture may be conditional on the shape, it should be \emph{independent of the pose}, so we encourage their disentanglement by adversarial means. 
This is necessitated by the unprojection step, which introduces a reliance of the inferred latent texture, $\widetilde{\xi}_T$, on the predicted pose, $\widetilde{E}$.
Hence, we train a small convolutional adversary network $A_R$ 
to infer the (best) rotation $\widetilde{R}_b$ 
from the (best) texture image $\widetilde{T}_{I,b}$; 
i.e., minimize $d_R(R_b, A_R(T_{I,b}))$.
Our model, $\fsi$ and $\fis$, is then trained to maximize the error of this regressor (only for the VC).
Separately, we additionally regularize the latent shape space $v$ with an SWD loss 
	(as used for $\xi_T$), 
	which helps the VC satisfy the shape critic $C_s$, 
	and is also used to help \textit{ex-post} distribution fitting to $v$
	(see Sec.\ \ref{sec:results:gen}).

\begin{figure}
	\centering
	\setlength{\tabcolsep}{1mm}
	\renewcommand{\arraystretch}{0.5} %
	\begin{tabular}{l|r}
		\includegraphics[width=0.22\textwidth]{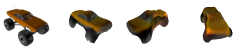} &
		\includegraphics[width=0.22\textwidth]{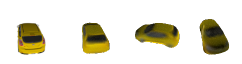} \\
		\includegraphics[width=0.23\textwidth]{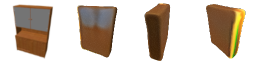} &
		\includegraphics[width=0.23\textwidth]{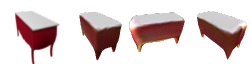}\\
		\includegraphics[width=0.23\textwidth]{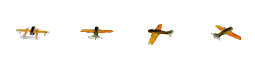} &
		\includegraphics[width=0.23\textwidth]{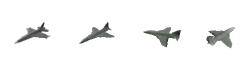}\\
		\includegraphics[width=0.23\textwidth]{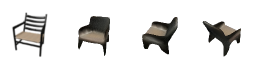} &
		\includegraphics[width=0.23\textwidth]{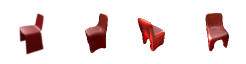}\\ 
		\includegraphics[width=0.23\textwidth]{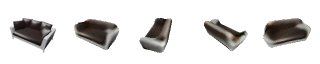} &
		\includegraphics[width=0.23\textwidth]{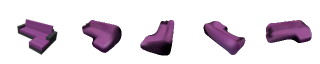}
	\end{tabular}

	\caption{Qualitative visualization of 3D reconstruction results. Per set: first image is the input $I$, second is the re-rendered reconstruction $\widehat{I}$, and the last two insets are different views of the output. Note that our masked input images are given a black background; hence the network chooses to fill ``holes'' with black (fourth row, left). 
	}
	\label{fig:qualrecon}
\end{figure}

\begin{table}
	\centering
	\setlength{\tabcolsep}{0.95mm}
	\renewcommand{\arraystretch}{1.1} %
	\small
	\begin{tabular}{@{\extracolsep{\fill}}lccccccccc@{\hspace{\tabcolsep}}}
		$\;$ & \multicolumn{3}{@{}c@{}}{F-score ($\tau$) $\uparrow$ } & \multicolumn{3}{@{}c@{}}{F-score ($2\tau$) $\uparrow$ } & \multicolumn{3}{@{}c@{}}{Chamfer $\downarrow$} \\
		\cmidrule(lr){2-4} \cmidrule(lr){5-7} \cmidrule(lr){8-10}
		Category  & 3DR & Ours & P2M & 3DR & Ours & P2M & 3DR & Ours & P2M \\
		\hline
		plane    & 41.5 & 70.0 & 71.1   & 63.2 & 81.9 & 81.4  & 0.90  & 0.37 & 0.48 \\ %
		cabinet & 49.9 & 46.7 & 60.4 & 64.8 & 60.6 & 77.2  & 0.74  & 0.90 & 0.38 \\ %
		car       & 37.8 & 58.2 & 67.9 & 54.8 & 70.7  & 84.2  & 0.85 & 0.56 & 0.27 \\ %
		chair    & 40.2 & 35.0 & 54.4 & 55.2 & 46.0 & 70.4  & 1.43  & 2.60 & 0.61 \\ %
		sofa     & 40.0 & 39.6 & 51.9 & 53.4 & 52.7  & 69.8  & 1.14  & 1.26 & 0.49 \\ %
		\hline
	\end{tabular}
	\vspace{0mm}
	\caption{ Reconstruction performance on the ShapeNet-based test set.
		Metrics: F-score (\%) at two thresholds, where $\tau=10^{-4}$ 
		and larger is better ($\uparrow$),
		 and Chamfer distance ($\times 1000$), 
		 where smaller is better ($\downarrow$). 
		Values for 3D-R2N2 (3DR) and Pixel2Mesh (P2M) are taken from \cite{wang2018pixel2mesh}. 
		While lower than the supervised counterparts, performance of our model still comes close, even with our more complex representational requirements; in particular, we are able to either exceed or come within 10\% of the F-score of 3D-R2N2 (see Sec.\ \ref{sec:results:recon}).
	}
	\label{table:3d_recon}
\end{table}

\begin{SCtable*}[12][h]
	\centering
	\setlength{\tabcolsep}{0.85mm}
	\renewcommand{\arraystretch}{1.0} %
	\small
	\begin{tabular}{@{}lcccccccccccc@{}}
		$\;$ & \multicolumn{4}{@{}c@{}}{IS $\uparrow$ } 
		& \multicolumn{4}{@{}c@{}}{FID $\downarrow$ } 
		& \multicolumn{4}{@{}c@{}}{KID $\downarrow$} \\
		\cmidrule(lr){2-5} \cmidrule(lr){6-9} \cmidrule(lr){10-13}
		Category   
		& VAE & Ours & OursU & GAN & 
		VAE & Ours & OursU & GAN &
		VAE & Ours & OursU & GAN  \\
		\hline
		plane    %
		& 2.36 & 2.98 & 2.86 & 3.39 & 
		92.3 & 58.9 & 59.4 & 11.5 &
		8.83 & 5.00 & 5.16 & 0.69  \\
		cabinet  %
		& 3.64 & 3.16 & 3.20 & 3.80 & 
		176.8 & 131.6 & 136.6 & 73.8 &
		15.73 & 10.94 & 11.55 & 6.53  \\
		car  %
		& 2.98 &  3.12 & 3.10 & 3.84 & 
		194.3 & 119.6 & 121.0 & 9.1 &
		18.61 & 10.37 & 10.51 & 0.491  \\
		chair    %
		& 4.10 & 3.96 & 3.90 & 5.21 & 
		119.9 & 108.8 & 110.8 & 45.8 &
		10.23 & 9.27 & 9.43 & 4.00  \\
		sofa      %
		& 3.73 & 3.86 & 3.79 & 4.66 & 
		171.9 & 94.3 & 98.1 & 27.7 &
		14.89 & 7.07 & 7.44 & 1.78  \\
		\hline
	\end{tabular}
	\caption{ 
		Performance on generative modelling.
		Metrics: 
			Inception Score (IS), 
			Frechet Inception Distance (FID), and 
			Kernel Inception Distance (KID; $\times 100$).
		OursU denotes \textit{un}conditional generation via \textit{ex-post} fitting.
		We perform similarly to the VAE, despite our constrained representation 
		(see Sec.\ \ref{sec:results:gen}).
	}
	\label{table:gen}
	 \end{SCtable*}

\section{Experimental Results}

For all experiments, 
	we use the shape data from ShapeNet \cite{chang2015shapenet} (without textures)
	and the rendered image data from 
	Choy et al.\ \cite{10.1007/978-3-319-46484-8_38}.
We remove internal mesh structure by voxelization and marching cubes \cite{lorensen1987marching}, using
Kaolin \cite{jatavallabhula2019kaolin}.
At no point is the relation between shapes and renders used by our model.
All models are implemented in PyTorch \cite{NEURIPS2019_9015}, 
and optimized with Adam \cite{kingma2014adam}. 
Hyper-parameters are tuned for visual quality on the training set, 
	and are the same across categories, except for the planes class (see Sec.\ \ref{suppsec:hyper}).
We refer the reader to the appendix for
additional experimental details (Sec.\ \ref{suppsec:experimental})
and results (Sec.\ \ref{suppsec:vis}).

\subsection{3D Reconstruction}
\label{sec:results:recon}
As a by-product of learning our weakly supervised modality translation model, 
we obtain the ability to perform 3D shape reconstruction from 2D images, 
via the first half of the VC (i.e., the image-to-shape function $\fis$).
Qualitative results are shown in Fig.\ \ref{fig:qualrecon}.
The network obtains the overall shape 
and captures visible appearance details fairly well (e.g., car windows).
However, it has difficulty texturing occluded parts %
(e.g., the backs of the chairs and cabinets).
Other works mitigate occlusion issues with symmetry constraints \cite{tulsiani2020implicit,kanazawa2018learning}, but these make assumptions on the object.
Issues with the fixed topology of the template are also visible in the chair reconstructions,
	as well as some of the planes (left inset).
Overall, however, the model is able to infer a reasonable shape and texture, despite the weak supervision.

For quantitative comparison (see Table \ref{table:3d_recon}), we use the same metrics and held-out testing set as Pixel2Mesh \cite{wang2018pixel2mesh}.
Since Pixel2Mesh is trained with the 3D ground truth in camera coordinates 
per input image, 
whereas our model decides on its own camera coordinates 
based only on explaining the image evidence, 
we align the 3D reconstructions in position and scale for fair comparison.
This is similar to a Procrustes analysis \cite{4767965} (centering and scalar scale matching), but without performing the rotation matching, since we expect our model to approximately output the correct orientation.
In particular, we sample points from our inferred mesh $\widetilde{M}_E$, transform it into camera coordinates, translate it (and the ground truth shape) to the origin based on the mean point value, and then scale our point set 
by an analytically computed scalar factor $s$, 
which corrects for the difference in 
(a) the distance to the camera along the optical axis and 
(b) the projective intrinsics of our renderer versus those used to render the ground truth data.
We note that this is important for quantitative evaluation, 
since point cloud metrics are sensitive 
to small differences in scale and rigid misalignments. 
As in \cite{wang2018pixel2mesh}, we utilize the Chamfer distance and F-score (precision and recall between point sets at a fixed threshold) to evaluate the method.

Results are shown in Table \ref{table:3d_recon}.
While our method does not perform as well as the Pixel2Mesh model, 
we are able to perform comparably to the voxel-based 3D-R2N2 model 
	(either exceeding its performance or coming within a 10\% margin by F-score), 
despite it being fully supervised with paired data.
Furthermore, we note that our model is less specialized than those we compare to, as it must learn additional capabilities (e.g., texture estimation, generative modelling); 
the capacity used by these additional tasks is expected to reduce performance 
(as is the weaker supervision).

\begin{figure}
	\centering 
	\includegraphics[width=0.49\textwidth]{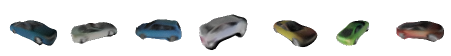}\\ %
	\includegraphics[width=0.49\textwidth]{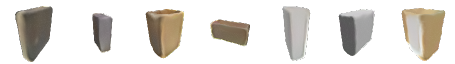}\\ %
	\includegraphics[width=0.49\textwidth]{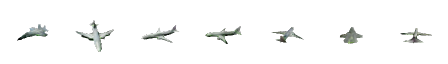}\\ %
	\includegraphics[width=0.49\textwidth]{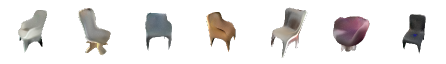}\\ %
	\includegraphics[width=0.49\textwidth]{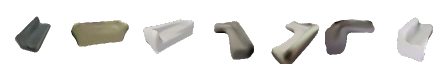} %
	\caption{Example generations from our shape-to-image function $\fsi$ with randomly sampled texture and pose.}
	\label{fig:qualgen}
\end{figure}

\subsection{Generative Modelling}
\label{sec:results:gen}

We next investigate the performance of our network as a generative model, 
	using the first half of the GC (via the shape-to-image function $\fsi$) to
	render an input shape $P$ based on a random pose $\xi_p$ and texture $\xi_T$.
Qualitative example generations are shown in Fig.\ \ref{fig:qualgen}. 	
We see that the model is able to output a diversity of shapes (especially in the chairs and sofas), 
	as well as acceptable textures (e.g., the details on the cars).
While point sets, $P$, are given as inputs, 
	we remark that fitting a surface mesh to a point cloud is itself a non-trivial problem 
	\cite{hanocka2020point2mesh},
	which our shape encoder $f_v$ must solve with the template, $S_T$, and input, $P$.
In addition, we note that white or grey monocoloured textures are disproportionately common, 
	particularly for the plane and sofa datasets, which is reflected in our generated distribution as well.
	
However, since our model is actually a \textit{conditional} generative model, %
we should remove this dependency for fairer comparison to methods
that  do not have access to a shape input.
Thus, to make our model \textit{un}conditional, 
we follow work in generative autoencoders
\cite{DBLP:journals/corr/abs-1903-12436,pmlr-v80-achlioptas18a}
and perform \textit{ex-post} density estimation to the regularized latent shape space $v$. 
In particular, we fit a Gaussian mixture model (GMM) 
$ \mathcal{Q}_v $
to the approximate aggregate posterior of the output of $f_v$.
In the context of generative models, this GMM serves as the latent prior over shapes. 
We therefore obtain an  unconditional model over textured shapes by sampling 
$v \sim \mathcal{Q}_v$ and $\xi_T, \xi_p\sim\mathcal{N}(0,I)$.

For comparison, we consider two standard generative models on 2D images: 
	a GAN \cite{goodfellow2014generative} and a VAE \cite{kingma2013auto,rezende2014stochastic}.
While a GAN is a useful gold-standard baseline, 
	we do not expect to outperform it on image generation, as it 
(i) does not have to learn an inference model, 
(ii) can manipulate pixels directly 
		(as opposed to colouring vertices, followed by rasterization),
and 
(iii) does not attempt to model 3D information.
An image VAE, on the other hand, has to (i) learn an encoder and (ii) ensure that the inferred latent variables are approximately Gaussian.
This is analogous to our model's texture representation, 
wherein the latent $\widetilde{\xi}_T$ is inferred from the input image and 
optimized to follow a Gaussian distribution.
Hence, 
	though it still has the advantage of operating on pixels and not having to extract or generate 3D shape and pose, 
an image VAE may be a more appropriate baseline than a GAN. 
For implementations, we use a WGAN-GP \cite{gulrajani2017improved} and a standard convolutional VAE, from public implementations in \cite{lee2020mimicry} and \cite{Subramanian2020}, respectively.
We show quantitative results with these models in Table \ref{table:gen}, 
using standard reference-free image quality metrics from the GAN literature:
Frechet Inception Distance (FID) \cite{heusel2017gans}, 
Inception Score (IS) \cite{salimans2016improved}, and 
Kernel Inception Distance (KID) \cite{binkowski2018demystifying}, 
computed via \cite{anton_obukhov_2020_3786540}.
We use the train-test split from \cite{wang2018pixel2mesh} for evaluation.
Unsurprisingly, the GAN scores the best in all cases, 
	but we find that our model generates images of similar quality as the VAE.
Our goal is not to show that there does not exist a VAE that could generate images of higher quality; rather, we wish to demonstrate that our network is able to perform at a similar level to a standard pixel-level generative model, despite the additional constraints on our model.
Finally, we remark that there is a fundamental domain gap between 
the output of our generative renderer 
(which has limited mesh resolution and a soft rasterization step) and 
that of the process that created the input pixel data, which limits our performance. %
For future work, we remark that texture quality could be improved by more sophisticated representations (e.g., texture fields \cite{oechsle2019texture}).

\begin{figure}
	\centering
	\includegraphics[width=0.23\textwidth]{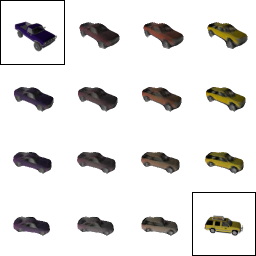}\hfill\vrule\hfill
	\includegraphics[width=0.23\textwidth]{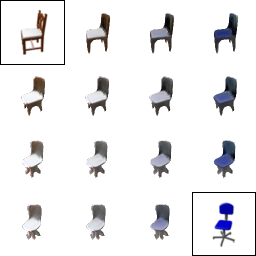} %
	\caption{Decoded renderings of latent space interpolations. Per-inset: boxed corner images represent start and end points (real images), vertical axis corresponds to interpolation in latent shape ($v$) and Euclidean pose ($E$), while the horizontal axis represents interpolation of the latent texture ($\xi_T$). Non-boxed opposing corners (upper-right and lower-left) may be viewed as \textit{texture transfers}. }
	\label{fig:interp}
\end{figure}

\begin{figure}
	\centering
	\begin{tikzpicture}[main node/.style={}]
	\node[yshift=0mm] (f1) {
		\includegraphics[width=0.47\textwidth]{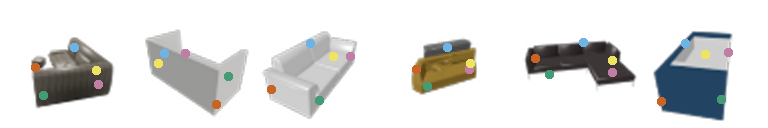}
	};
	\node[below=0mm of f1,yshift=3.5mm] (f2) {
		\includegraphics[width=0.47\textwidth]{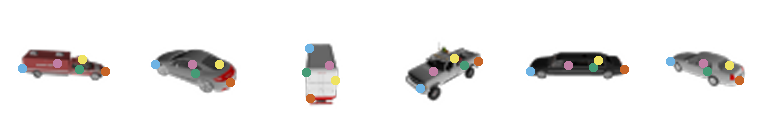}
	};
	\node[below=0mm of f2,yshift=5mm] (f3) {
		\includegraphics[width=0.47\textwidth]{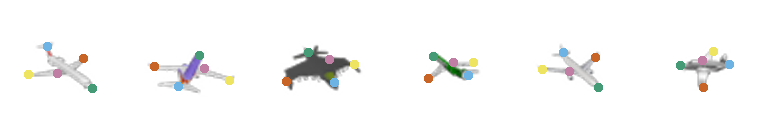}
	};
	\draw[-latex,black,thick,-] (-3.9,-0.7) -- (-2.9,-0.7);
	\draw[-latex,black,thick,-] (-3.9,-1.7) -- (-2.9,-1.7);
	\draw[-latex,black,thick,-] (-1.15,-2.9) -- (-0.15,-2.9);
	\end{tikzpicture}
	\caption{Unsupervised corresponding keypoint identification. Each row shows a set of real input images from our dataset, along with several marked keypoints in correspondence across images (obtained by projecting the inferred output shape onto each one). The same colour of point denotes the same template vertex index across images. Note that we show the keypoint even if the originating vertex is occluded in the image. 
		Underlines highlight shapes that the model incorrectly inferred ``backwards''.
	}
	\label{fig:corres}
\end{figure}
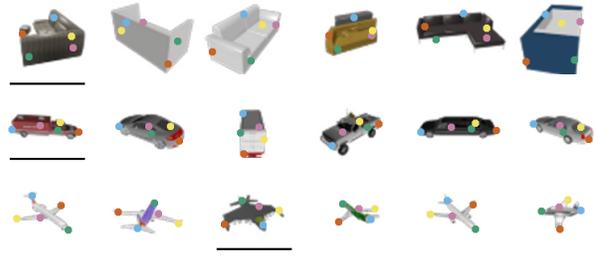

\subsection{Latent Textured Shape Representation}

We highlight two additional applications of our network to representation learning: 
(i) latent shape-pose-texture disentanglement
and 
(ii) unsupervised correspondence discovery.
The first task concerns the interpretable nature of our latent ``graphics code'' representation, showing that we can manipulate the 3D pose, intrinsic shape, and texture of an object, independently.
This is showcased in Fig.\ \ref{fig:interp}, where we interpolate through latent shape and pose, separately from texture. 
The second task considers the natural geometric correspondence that occurs 
	due to the use of a canonical template, $S_T$,
	which enables the network to assign a meaning to each vertex 
	that is consistent across shapes. %
The order dependent cyclic shape loss in the GC ($M$-to-$\widehat{M}$) further encourages consistency,
	as it demands inference of the \textit{same} position for each  vertex, across the rendering process. %
Thus, by looking at the projected positions of specific template nodes (i.e., inferring $\widetilde{M}_E$ from the input, and marking where the chosen vertices project to in $I$), we obtain approximately corresponding points on the original images, similar to an unsupervised keypoint discovery method.
 We show  examples of this in Fig.\ \ref{fig:corres}, where each colour point corresponds to the \textit{same} template node index.
The results show that the network is often able to identify ``corresponding'' points across images, despite changes in pose, shape, and appearance. They also reveal areas of difficulty for the network (e.g., confusing the front and back of approximately symmetric objects, as in the underlined insets).

\section{Conclusion}

We have presented a unified learning framework for \textit{modality translation} 
	between 2D images and 3D textured shape.  
Our method combines a 
shape-to-image mapping, $\fsi$, which performs generative rendering, with an
image-to-shape mapping, $\fis$, that infers a latent 3D representation from a single image.
We learn these functions by optimizing over two cyclic consistency losses that alternate these mappings: the \textit{graphics} and \textit{vision} cycles.
Our model works directly on triangle meshes, and is trained with weak supervision using \textit{unpaired} data.
We demonstrated its potential on single-image 3D reconstruction, textured shape generation, and latent representation learning.
However, our method currently does not handle images with backgrounds, is topologically constrained in 3D shape by its use of a deformable template based on nodal offsets (e.g., Fig.\ \ref{fig:interp}, right inset), 
and does not separate lighting from texture. 
Nevertheless, we expect our model to be useful in helping machine learning models make better use of data with less supervision, as well as in obtaining generative representations with interpretable 3D latent factors.

{\small
	\bibliographystyle{ieee}
	\bibliography{defs,b}
}

\appendix

\section{Model Details}
\label{suppsec:model}

We first provide details on the architectures of the component networks defining our modality translation functions. %
We set 
the number of pose hypotheses to
$n_h = 4$,
and the various encoding dimensionalities as
$\dim (v) = 64$,
$\dim (q) = 512$,
$\dim (\xi_p) = 16$, 
and
$\dim (\xi_T) = 128$.

Notationally, we write \verb|LBA(n1,n2,...)| to mean a multilayer perceptron with hidden layers of size \verb|n1,n2,...|, ending in a linear layer.
Each layer consists of a fully connected linear layer ($\verb|L|$), 
followed by batch normalization (\verb|B|) \cite{ioffe2015batch}, and 
non-linear ELU activation (\verb|A|) \cite{clevert2015fast}.
We append a \verb|BA| to the end, if we include a \verb|B| and \verb|A| after the final linear layer.
We also use the notation $f : a \rightarrow b$ to remind the reader of the input and output variables of the mapping $f$. 
The same variable definitions are used as in the main text.

Model component architectures are as follows:
\begin{itemize}	
	\item 
	The pose decoder $f_p : \xi_p \rightarrow (r,t)$ is given by \verb|LBA(64)|, but note the additional step converting from Tait-Bryan angles $r\in\mathbb{R}^3$ to a rotation matrix $R\in SO(3)$, so that the Euclidean transform for rigid pose is given by $E=(R,t)$.
	
	\item 
	The shape encoder $f_v : P \rightarrow v$ is implemented as a standard PointNet \cite{qi2017pointnet}, without spatial transformers. Hidden layers have sizes 
	\verb|64, 128, 1024, 512,| and \verb|512|.
	
	\item 
	The nodal perturbation decoder $f_\delta : v \rightarrow \delta$ is written
	\verb|LBA(600,900,1200)|.
	
	\item 
	Template mesh $S_T$: we use a sphere with $|\mathcal{V}| = 1002$ and $|\mathcal{F}| = 2000$.
	
	\item 
	The image encoder $h_q : I \rightarrow q $ is a standard ResNet-18 backbone \cite{he2016deep}.
	
	\item 
	The latent shape inferrer $h_v : q \rightarrow \widetilde{v}$ is a single fully connected layer.
	
	\item 
	The pose inferrer $h_p : q \rightarrow\widetilde{E}$ is given by \verb|LBA(12*n_h)BA|, but note the additional step converting from Tait-Bryan angles to a rotation matrix, and the presence of hypotheses.
	
	\item 
	The pose hypothesis probability estimator $h_e : q \rightarrow H$, where $H = (p_1,\ldots,p_{n_h})$, is given by \verb|LBA(32*n_h)-softmax|. 
	
	\item 
	The latent texture inference network $h_T : (q, \widetilde{v}, U) \rightarrow \widetilde{\xi_T}$ is decomposed into three subnetworks:
	\begin{enumerate}
		\item
		A latent shape preprocessor 
		$g_v : v \rightarrow z_v$, where $\dim(z_v) = 32$, is given by \verb|LBA(128)BA|.
		\item
		An unprojection preprocessor
		$g_u : U \rightarrow z_u$, where $ \dim(z_u) = 512 $, is given by \verb|LBA(512)BA|.
		\item 
		The final sub-network for latent texture inference,
		$g_z : (q, z_v, z_u) \rightarrow  \widetilde{\xi}_T$, is written \verb|LBA(3*Dz,2*Dz)|, 
		where \verb|Dz=32+512+512|.
	\end{enumerate}
	
	\item 
	The texture decoder $f_T : (\xi_T, M) \rightarrow T$ consists of three sub-operations:
	\begin{enumerate}
		\item 
		The mesh preprocessor
		$g_M : M \rightarrow z_M$, where $\dim(z_M) = 400$, is written \verb|LBA(800)|.
		\item 
		The texture image generator
		$G_{T_I} : (\xi_T, z_M) \rightarrow T_I$ is a standard image generation network \cite{gulrajani2017improved}, utilizing transposed convolutions in residual blocks. 
		Specifically, it consists of a linear layer, (to dimensionality $4^2 \times 1024$, followed by reshaping into a $4\times 4$ image. 
		Four residual generator blocks are then applied, 
		which halve the channel dimension and double the spatial resolution,
		and a final 2D conv layer (after a BatchNorm and ReLU) is applied, 
		followed by tanh.
		The texture image $T_I$ is a $64\times 64$ RGB image.
		(See implementation in \cite{lee2020mimicry}, under $64\times 64$ WGAN generators).
		\item 
		We finally obtain the nodal colours from the texture image via bilinearly interpolated grid sampling
		\cite{jaderberg2015spatial}.
	\end{enumerate}
	
	\item 
	The SoftRas differentiable renderer \cite{liu2019soft} is used for $\mathcal{R} : (T, M_E) \rightarrow I$.
	We set the probabilistic spatial sharpness to $\sigma = 10^{-5}$ and the colour aggregation sharpness to $\gamma = 10^{-4}$.
	
\end{itemize}

\subsection{Detachments}
We found it useful to make several judicious detachments within the computation graph (i.e., prevent gradient backpropagation flow along particular connections). This includes the connection between $M$ and $f_T$, $v$ and $h_T$, and $E$ and $f_E$.
The former two prevent the network from using node positions to store texture information, while the latter mitigates instability from the adversarial gradients flowing into the pose generator. 

\subsection{Timing}

We note that a single complete training iteration for our model takes ${\sim }3.3$s, 
including forward and backward passes for both cycles (graphics and vision), 
as well as for all critics and adversaries.
Our model is trained for $10^5$ iterations (see Sec.\ \ref{sec:supp:curriculum} for training regime details), 
which takes ${\sim}90$ hours on a single NVIDIA Tesla V100 GPU.
The vision cycle forward pass costs ${\sim}1.2$s, 
while that of graphics cycle costs ${\sim}0.9$s. 
The costliest critics are the 
Sinkhorn-based pose matcher (${\sim }0.1$s) 
and the generated render adversary $C_I$ (${\sim }0.09$s).
For the translations alone, running 
shape-to-image translation takes ${\sim}0.25$s,
while image-to-shape translation takes ${\sim}0.4$s.
These numbers utilized a single NVIDIA Tesla V100 GPU and a batch size of 64.

\section{Loss Details}
\label{suppsec:losses}

We now expand on the loss terms described in the main text of the paper.
Recall that the complete objective function is given by
\begin{equation}
\mathcal{L} = \mathcal{L}_G + \mathcal{D}_G + \mathcal{L}_V + \mathcal{D}_V +  \mathcal{L}_R.
\end{equation}

\subsection{Graphics Cycle Losses}

\subsubsection{Distribution Matching Loss $ \boldsymbol{\mathcal{D}}_{\boldsymbol{G}} $}%

The distribution matching loss for the graphics cycle, $\mathcal{D}_G$, is written as
\begin{equation}
\mathcal{D}_G = - \gamma_{g,I} C_I(\widetilde{I}) - \gamma_{g,T} C_T(\widetilde{T}_I) + 
\gamma_{g,E} \mathcal{S}_D[ \mathcal{B}_{\widetilde{E}} \mid\mid E ],
\end{equation}
where the first two terms (using $C_I$ and $C_T$) are WGAN generator losses \cite{arjovsky2017wasserstein,gulrajani2017improved} (see Sec.\ \ref{sec:criticdetails}), 
and the final term is the Sinkhorn distance \cite{cuturi2013sinkhorn} between the pose buffer $\mathcal{B}_{\widetilde{E}}$ of recently inferred (best hypothesis) poses from the vision cycle (buffer size 128) and the currently sampled batch of poses, simply denoted here as $E = (R,t)$. This latter term is decomposed into two terms, for the rotations and translations, respectively:
\begin{equation}
\mathcal{S}_D[ \mathcal{B}_{\widetilde{E}} \mid\mid E ]
= \mathcal{S}_D[ \mathcal{B}_{\widetilde{R}} \mid\mid R ]
+  \mathcal{S}_D[ \mathcal{B}_{\widetilde{t}} \mid\mid t ].
\end{equation}
We use the $L_2$ ground metric for $t$ and the geodesic distance on $SO(3)$, denoted $d_g$, for $R$:
\begin{equation}
d_R(R_1, R_2) = \arccos( [\text{tr}(R_1 R_2^T) - 1] / 2 ).
\end{equation}
We set the weights as
$  \gamma_{g,I}  = 0.2 $,
$  \gamma_{g,T} = 0.2 $, and
$  \gamma_{g,E} = 10 $.

\subsubsection{Cyclic Consistency Loss $ \boldsymbol{\mathcal{L}}_{\boldsymbol{G}} $}%

Recall that the template mesh is given by $S_T = (\mathcal{V},\mathcal{E},\mathcal{F})$. 
Let $M = (\mathcal{V}_M,\mathcal{E},\mathcal{F})$ be the canonical deformation of $S_T$,
which alters neither the edges nor the faces,  
where $\mathcal{V}_M \in \mathbb{R}^{|\mathcal{V}|\times 3}$ 
is the matrix of nodal coordinates.
The cyclic consistency loss for the graphics cycle, $\mathcal{L}_G$, is then written as
\begin{align}
\mathcal{L}_G &= \frac{\gamma_{g,v}}{\dim(v)} || v - \widehat{v} ||_1   +  
\frac{\gamma_{g,M}}{|\mathcal{V}|} || \mathcal{V}_M - \widehat{\mathcal{V}}_M ||_F^2 \; + \\
& \;\;\;\;\;
\frac{\gamma_{g,t}}{3} || t - \widehat{t} ||_2^2  +  
\frac{\gamma_{g,\xi_{T}} }{\dim(\xi_T)} || \xi_{T} - \widehat{\xi}_T ||_2^2 \; + \\
& \;\;\;\;\;
\frac{\gamma_{g,R}}{\pi} \sum_{i=1}^{n_h} \widehat{p}_i d_R( R, \widehat{R}_i ) + %
\mathscr{D}(P,M),
\end{align}
where the input oriented point set $P=(X,N)$ is written as a tuple of 3D point coordinates $X$ and normals $N$. 
The final term, matching the inferred canonical template mesh $M$ to the input shape $P$ is written as
\begin{equation}
\mathscr{D}(P,M) = \gamma_{g,X} D_C(X_M,X) - \gamma_{g,N} | N \cdot N_M |,
\end{equation}
where $X_M$ and $N_M$ are uniformly sampled from $M$, and $D_C$ is the Chamfer distance.
We denote $| N \cdot N_M |$ to mean the average per-point dot product 
between a surface normal $n\in N$ at a point $x \in X$, 
and the surface normal $n_M \in N_M$, associated to the point $x_M \in X_M$ that is closest to $x$.
In other words, the first term matches the point set and the second matches the normals,
as in \cite{gkioxari2019mesh}. 
We perform this loss on the vertices, as well as the samples, to ensure each template vertex is given a signal at every step.
We further remark that $\widehat{p}$ refers to the pose hypothesis probabilities computed when applying the modality translation $f_{I\rightarrow S}$, in the second half of the graphics cycle.

We set the weights as
$\gamma_{g,v} = 5 $,
$\gamma_{g,M} = 1000 $,
$\gamma_{g,t} = 10 $,
$\gamma_{g,\xi_T} = 5 $,
$\gamma_{g,R} = 5 $,
$\gamma_{g,X} = 2000 $, and
$\gamma_{g,N} = 5 $.

\subsection{Vision Cycle Losses}

\subsubsection{Distribution Matching Loss $ \boldsymbol{\mathcal{D}}_{\boldsymbol{V}} $}%

The distribution matching loss for the vision cycle, $\mathcal{D}_V$, consists of two loss terms:
\begin{equation}
\mathcal{D}_V = -\gamma_{v,S} C_s(\widetilde{v},\widetilde{M}) + 
\gamma_{v,\xi_T} \mathfrak{D}_{\text{SWD}}[ \widetilde{\xi}_T \mid\mid \xi_T ], 
\end{equation}
where $C_s$ is the WGAN shape critic loss (see Sec.\ \ref{sec:criticdetails}) and $\mathfrak{D}_{\text{SWD}}$ is the sliced Wasserstein distance (SWD) \cite{deshpande2018generative,rabin2011wasserstein} loss between the inferred $\widetilde{\xi}_T$
and the normally distributed latent textures $\xi_T \sim \mathcal{N}(0,I)$ (using 150 projections).
We remark that the loss on $ \widetilde{\xi}_T $ is only applied to those derived from the best hypothesis for a given input.
We set the weights as $ \gamma_{v,S} = 1 $ and $ \gamma_{v,\xi_T} = 5$.

\subsubsection{Cyclic Consistency Loss $ \boldsymbol{\mathcal{L}}_{\boldsymbol{V}} $}%

The cyclic consistency loss, $\mathcal{L}_V$, for the vision cycle only matches the input image, $I$, and its rendered reconstruction, $\widehat{I}$, via pixelwise and perceptual metrics:
\begin{equation}
\mathcal{L}_V = \sum_{i=1}^{n_h} p_i
\left[
\frac{ \gamma_{v,I} }{4N_p} 
|| I - \widehat{I}_i ||_1
+
\frac{ \gamma_{v,p} }{N_b}	
\sum_{j=1}^{N_b} \frac{|| c_{i,j} - \widehat{c}_{i,j} ||_1}{B_j N_{p,j}} 
\right],  %
\end{equation}
where 
the first term is the pixelwise $L_1$ loss 
(normalized by the number of pixels, $ N_p $, and channels)
and
the second term is the perceptual loss over critic layers
(normalized by the the channel dimensionality of block $j$, denoted $B_j$, 
and number of spatial features, $N_{p,j}$),
both weighted by hypothesis probability $p_i$.
We note that $N_b = 5$ is the number of discriminator block outputs we utilize 
(one per residual block in $C_I$), while
$c_{i,j}$ and $\widehat{c}_{i,j}$ are the feature maps 
from critic block $j$ and hypothesis $i$, 
for $I$ and $\widehat{I}_i$, respectively. 
This perceptual loss is quite similar to those used in \cite{larsen2016autoencoding,wang2018high}.
We set the weights as $\gamma_{v,I} = 40$ and $\gamma_{v,p} = 200$.

\subsection{Regularization Losses}

The final loss term is composed of an assortment of \textit{regularization objectives}:
\begin{align}
\mathcal{L}_R &= \gamma_{r,m} [ L_m(M) + L_m(\widetilde{M}) ] \; + \\
& \;\;\;\;\;
\gamma_{r,\delta} [ L_\delta(\delta) + L_\delta(\widetilde{\delta}) ] \; + \\
& \;\;\;\;\;
L_v(v,\widetilde{v})  + 
L_D(\widetilde{\xi}_T, \widetilde{R}, \widetilde{t}) + 
\gamma_{r,A} L_A(\widetilde{T}_I, \widetilde{R}),
\end{align}
where 
$L_m$ is a mesh regularization loss (for preventing pathological geometry),
$L_\delta$ promotes smoothness in the nodal offset vector (similar in purpose to $L_m$),
$L_v$ regularizes $v$ to be approximately Gaussian 
(to assist in our \textit{ex-post} density estimation procedure for removing shape-conditioning in the generative process),
$L_D$ regularizes the pose hypothesis distributions 
(as well as the latent texture ones),
and 
$L_A$ denotes the adversarial texture-pose disentanglement loss 
(which reduces the dependence of the texture on the pose).
We set the weights
$\gamma_{r,m} = 0.05$, 
$\gamma_{r,\delta} = 0.05$, and
$\gamma_{r,A} = 0.35$.

We next expand on these terms in detail. 
Recall that $M = (\mathcal{V}_M,\mathcal{E},\mathcal{F})$ is a nodal deformation of the template mesh. %
The mesh regularizer borrows three terms from the literature:
\begin{align}
L_m(M) &= 
\frac{\gamma_{r,m,e}}{|\mathcal{E}|} \sum_{(v_1, v_2) \in \mathcal{E}} || v_1 - v_2 ||^2_2
\;+ \\
& \;\;\;\;\;
\gamma_{r,m,\ell} L_\text{Lap}(M)
+
\gamma_{r,m,f} L_\text{Flat}(M),
\end{align}
where 
the first term regularizes the edge lengths \cite{wang2018pixel2mesh,gkioxari2019mesh},
the second is a Laplacian loss \cite{DBLP:journals/corr/abs-1901-05567} 
(designed to prevent excessive curvature),
and the third is a flatness loss \cite{DBLP:journals/corr/abs-1901-05567} 
(which encourages local smoothness of the surface normals). 
We set the weights
$\gamma_{r,m,e} = 1500$,
$\gamma_{r,m,\ell} = 10$, and
$\gamma_{r,m,f} = 1$.

The next term regularizes the smoothness of the nodal perturbation directly:
\begin{equation}
L_\delta(\delta) = 
\frac{1}{|\mathcal{F}|} \sum_{f\in\mathcal{F}} \sum_{d = 1}^3
\mathbb{V}[\delta[f]]_d,
\end{equation}
where $\mathbb{V}[\delta[f]]_d$ is the empirical variance of $\delta$ values over triangle face 
$ f\in\mathcal{F}$, for spatial dimension $d$.

The regularizer on $v$ is simply an SWD loss on the latent shape 
(similar to the one used for $\widetilde{\xi}_T$):
\begin{equation}
L_v(v,\widetilde{v}) = 
\gamma_{r,v} \, \mathfrak{D}_{\text{SWD}}[ v \mid\mid \eta ]
+
\gamma_{r,\widetilde{v}} \, \mathfrak{D}_{\text{SWD}}[ \widetilde{v} \mid\mid \eta ],
\end{equation} 
where $\eta \sim\mathcal{N}(0,I)$, and we set 
$ \gamma_{r,v} = 5$
and 
$ \gamma_{r,\widetilde{v}} = 0.1$ 
(because the shape critic $C_s$ regularizes $\widetilde{v}$ as well).

We next consider the ``diversification'' losses on the pose hypotheses, as well as on the resulting texture hypotheses:
\begin{align}
L_D(\widetilde{\xi}_T, \widetilde{R}, \widetilde{t})
&=
- 
\frac{\gamma_{r,d,R}}{\pi} 
\sum_{i=1}^{n_h}
\sum_{j=1}^{n_h} p_i p_j d_R(\widetilde{R}_i, \widetilde{R}_j) \; + \\
& \;\;\;\;\;
\frac{
	\gamma_{r,d,\xi_T} }{n_h \dim(\xi_T)}
\sum_{k=1}^{n_h} || \widetilde{\xi}_{T,k} - \widetilde{\xi}_{T,b} ||_2^2 \; + \\
& \;\;\;\;\;
\frac{
	\gamma_{r,d,t} }{ 3 n_h }
\sum_{k=1}^{n_h} || \widetilde{t}_{k} - \widetilde{t}_{b} ||_2^2,
\end{align}
where the three terms encourage diversity in the rotations $\widetilde{R}$, 
and discourage it for $\widetilde{\xi}_T$ and $\widetilde{t}$.
The index $b$ refers to the ``best'' hypothesis (i.e., $\arg\max_j p_j$).
We set the weights
$ \gamma_{r,d,R} = 0.1$,
$ \gamma_{r,d,\xi_T} = 1 $, and
$ \gamma_{r,d,t} = 10 $.
We remark that we are trying to encourage diversity in the rotations
(as in \cite{kulkarni2019canonical}) in an effort to spur exploration, but found reducing it for the translations helped with numerical stability 
(whereas, for the latent shape, we desire the same texture regardless of pose).

The final regularization term is designed to encourage disentanglement of rigid pose and texture.
We trained an adversary network, $A_R$, which attempted to predict the rotation component $\widetilde{R}$ of our inferred Euclidean transform $\widetilde{E}$, from the texture image $\widetilde{T}_I$. 
Note that $\widetilde{T}_I$ depends directly on $\widetilde{R}$ through the unprojection.
We implemented $A_R$ as a ResNet-20 \cite{he2016deep}, which regressed a quaternion from the image, followed by conversion to a rotation matrix.
We trained $A_R$ with one gradient descent step per generator update (i.e., for the main model), 
using Adam \cite{kingma2014adam} (learning rate 0.0005 and the same $\beta$ values as for the main model).
Adversary training is done via minimization of $ d_R(\widetilde{R}, A_R(\widetilde{T}_I)) $.
Hence, our main model's regularization loss is given by
\begin{equation}
L_A(\widetilde{T}_I, \widetilde{R}) = - d_R(\widetilde{R}_b, A_R(\widetilde{T}_{I,b})),
\end{equation}
where we note that this loss is only applied to the \textit{best} hypothesis (i.e., with highest estimated probability), denoted with index $b$.

\subsection{Critic Details}
\label{sec:criticdetails}
We next discuss the details of the critic architectures. 
All critics were optimized with Adam \cite{kingma2014adam}
with a learning rate of 0.0001,
used the WGAN-GP loss formulation \cite{arjovsky2017wasserstein, gulrajani2017improved} with
a ``drift'' penalty \cite{karras2018progressive} on the mean squared output value,
and trained with one descent step per generator update. 
Gradient and drift penalty weights were set to 10 and 0.01, respectively.
An $L_2$ weight decay of $10^{-3}$ was also applied to all critics, as in \cite{gulrajani2017improved}. All critics have a scalar output.

The details per critic model are given as follows:

\begin{itemize}
	
	\item 
	The critic on generated images (i.e., sampled renders), $C_I(\widetilde{I})$, 
	is given by a standard WGAN-GP convolutional discriminator composed of residual blocks: 
	an initial block without downsampling, followed by four blocks with downsampling, and
	then a ReLU, mean pooling, and a final linear layer. Hidden channel dimensionalities go from 64 to 1024 in factors of two.
	We reiterate that the use of a \textit{soft} rasterizer \cite{liu2019soft} and mesh of limited resolution 
	(with interpolated per-triangle colouring),
	in addition to a weaker lighting model,
	leads to a fundamental domain gap compared to real images.
	Hence, we train the image critic $C_I$ with a mix of real images and inferred image reconstructions: 
	with probability $\rho_r = 0.5$, we replace an image $I$ with its vision cycle (VC) reconstruction $\widehat{I}$.
	This is designed to increase distributional overlap, 
	which is known to stabilize adversarial training (e.g., \cite{8953819}).
	
	\item 
	The critic on generated texture images, $C_T(\widetilde{T}_I)$, employs the same architecture as that of $C_I$. However,  it is trained with the texture images from the vision cycle (as we do not assume access to ground truth textures).
	
	\item 
	The critic on inferred shapes, 
	$C_s(\widetilde{v},\widetilde{M})$, 
	uses two networks. 
	The first processes $\widetilde{M}$ to a 128D vector $\widetilde{y}_M$, 
	and is given by \verb|LBA(512)BA|.
	The second is a vector critic on the concatenation $(\widetilde{v},\widetilde{y}_M)$, 
	with architecture \verb|LBA(512,256,128)|.
	
\end{itemize}

\section{Experimental Details}
\label{suppsec:experimental}

\subsection{Training Details}

All models were implemented in 
PyTorch \cite{NEURIPS2019_9015}, 
and optimized with 
Adam \cite{kingma2014adam} ($\beta_1 = 0.5, \beta_2=0.99$).
The learning rate was fixed to $0.0004$, using a batch size of 64.
To provide normalized input, we scaled all input shapes $S$ into the unit bounding box, 
and took 1000 points (with normals) per data shape. %
When computing losses, we also sampled 1000 points from our template.

\subsubsection{Curriculum Learning}
\label{sec:supp:curriculum}

To encourage stable training, we follow a learning curriculum for our training regime, inspired by techniques in domain randomization \cite{tobin2017domain,tremblay2018training}.
In the first stage, we train the graphics cycle (GC) mapping from the input shape, $P$, to the canonically deformed template, $M$, for $n_{T_1}$ iterations.
We then define a ``domain randomized'' version of the GC (abbreviated DR-GC), where Euclidean pose $E$ and texture $T$ are randomly sampled from hand-crafted latent distributions, rather than via the learned densities defined by $\xi_p$ and $\xi_T$, respectively.
The pose distribution is chosen to simulate a camera sampled from the upper-hemisphere of the object, along with a random displacement in a box around the origin.
The texture distribution is formed by choosing a random 3D plane in space and colouring the vertices on each side of the plane a constant, randomly chosen colour. 
The latent texture reconstruction objective is replaced by a loss 
between the textures $T$ and $\widehat{T}$ instead.
We train with this DR-GC for $n_{T_2}$ iterations. 
We then introduce the VC and train it with the DR-GC for $n_{T_3}$ iterations, 
followed by introducing the normal GC (with both VC and DR-GC) for $n_{T_4}$ iterations. 
For this fourth stage, we linearly anneal in the losses from the GC, while annealing out the losses from the DR-GC at the same time.
The final stage utilizes the full GC and VC for the remaining $n_{T_5}$ iterations.
We set 
$n_{T_1} = 5000$,
$n_{T_2} = 3000$,
$n_{T_3} = 3000$,
$n_{T_4} = 3000$, and 
$n_{T_5} = 10^5 - \sum_{i = 1}^4 n_{T_i}$.

\subsubsection{Hyper-Parameter Details}
\label{suppsec:hyper}

All hyper-parameters were chosen for visual quality, and the same across all categories, 
except for planes (see Sec.\ \ref{suppsec:model} and \ref{suppsec:losses}).
For the planes category, 
we found that the model struggled to generate images of equivalent quality to the other categories, 
for both the vision and graphics cycles.
We speculate this is potentially due to the smaller size of the objects relative to the image, 
as well as the degenerate cases induced when viewing planes from the side.

We made two major changes to mitigate such issues, specific to the planes category.
First, due to the relative conformity of plane shapes, 
we found that utilizing a specialized template was helpful.
We used a plane model from the ShapeNet training set, and computed a UV parametrization via boundary first flattening 
(with 16 auto-placed cones) \cite{sawhney2017boundary}.
Unlike for other categories, we let the template vertex positions be updated during learning as well.
Second, we altered several of the hyper-parameters, as follows:
$\gamma_{v,I} = 100$,
$\gamma_{v,p} = 100$,
$\gamma_{r,v} = 0.5$,
$\gamma_{r,a} = 0$,
$\text{dim}(v) = 128$,
$\gamma_{g,v}$,
$\gamma_{g,I} = 0.5$,
$\gamma_{g,R} = 1$,
$\gamma_{g,\xi_{T}} = 1$,
$ \gamma_{g,\text{TR}} =  10$,
$\gamma_{g,T} = 1$,
$\gamma_{v,\xi_T} = 1$,
$\gamma_{r,m} = 0.005$,
and $\gamma_{r,\delta} = 0.1$.
Note that $\gamma_{g,\text{TR}}$ is the weight coefficient on a nodal texture reconstruction loss term in the graphics cycle,
\begin{equation}
L_{g,\text{TR}} = \frac{\gamma_{g,\text{TR}}}{3 n_h |\mathcal{V}|}\sum_{i = 1}^{n_h} || T - \widehat{T}_i ||_1,
\end{equation} 
which was not used for the other categories.
The plane model with these altered hyper-parameters is used for all results and images. %

For completeness, we also record the quantitative evaluations of the \textit{unmodified} planes model 
(i.e., the one having identical hyper-parameters to the other categories): 
for reconstruction, 
F-score ($\tau$), F-score ($2\tau$), and Chamfer were 63.2, 75.8, and 0.55 
(Table \ref{table:3d_recon});
for conditional generation,
IS, FID, and KID were 2.69, 65.9, and 6.11 (see Table \ref{table:gen}); 
and 
for unconditional generation,
IS, FID, and KID were 2.65, 67.8, and 6.38  (see Table \ref{table:gen}).

For all hyper-parameters, we expect that additional experimentation (including category-specific hyper-parameter search) would further improve results in general.

\subsection{Dataset Details}

We show the number of data points (images and shapes) in the following table:

\begin{center}
	\setlength{\tabcolsep}{4pt}
	\begin{tabular}{c|ccccc}
		Dataset 						& Car          & Plane      & Sofa         & Chair       & Cabinet   \\\hline
		Train-S$^\text{SN}$ 	 & 5997        &  3236     &  2539       &  5423      &  1258       \\
		Train-I 	 					  & 143928    &  77664    &  60936    &  130152   &   30192   \\
		Test-S$^\text{P2M}$     & 7496       &   4045     &  3173       &   6778      &   1572      \\
		Test-I  	 					  & 36000	  &  19416     & 15240      &   32352   &   7560   
	\end{tabular}
\end{center}
The appended ``-S'' and ``-I'' refer to 3D shapes and 2D images, respectively.
The SN and P2M superscripts refer to shapes drawn 
from ShapeNet (SN) itself versus the Pixel2Mesh (P2M) 3D models data.
Note that these are different because ShapeNet shapes are given in pose-aligned form 
(canonical object space), 
whereas the data for Pixel2Mesh uses camera-space shape models 
(with multiple poses, and thus meshes, per ShapeNet shape).

\subsection{3D Reconstruction}

We use the same testing set as \cite{wang2018pixel2mesh} for fair comparison (as well as the same evaluation metrics).
For a given ground truth image and associated scene-space shape 
$(I_\text{gt}, P_\text{gt})$, 
we run $f_{I\rightarrow S}(I)$ to obtain our Euclidean transformed inferred shape, $\widetilde{M}_E$.
We evaluate reconstruction accuracy by 
(i) translating and scaling our output mesh to match $P_\text{gt}$,
(ii) uniformly sampling points from the surface of $\widetilde{M}_E$ equal to the number of points in $P_\text{gt}$,
and 
(iii) computing Chamfer distance and F-score between our output and the ground truth shape 
using the Kaolin library \cite{jatavallabhula2019kaolin}.

For (i), we first mean-centre the mesh. 
Then, we approximately correct for the scales induced by the differing camera coordinates. 
Let $d_z$ and $f_\ell$ be the distance along the optical axis from the camera (pre-centring) and the focal length of the camera, respectively, {for the ground-truth data}. Let $\widehat{d}_z$ and $\widehat{f}_\ell$ be the same for our output (pre-centring). Then our scaling factor is written
\vspace*{-2mm}
\begin{equation} %
s = \frac{d_z}{\widehat{d}_z} \frac{\widehat{f}_\ell}{f_\ell}. \vspace*{-1mm}
\end{equation}
We remark here that, due to the weak supervision, 
our model has weak constraints in terms of scale and distance from the camera; for instance, a slightly smaller shape $\widetilde{M}$, but closer to the camera (via $\widetilde{t}_z$), may still produce a valid solution (in terms of the VC losses).
One thus might expect an additional scalar scaling factor is needed to correct for this (e.g., based on the variance of the point set, as often used in Procrustes analyses \cite{4767965}). 
However, our results did not utilize one, as it reduced performance compared to just using $s$ alone.

\subsection{Generative Modelling}

We next describe the generative modelling experiments in greater detail.
In all cases, we used the rendered data from \cite{10.1007/978-3-319-46484-8_38}, and the train-test split from \cite{wang2018pixel2mesh}. 
Any render from a shape in the test set was placed in the test set for the renders; 
thus, our evaluation 
(using Inception Score \cite{salimans2016improved}, 
Frechet Inception Distance \cite{heusel2017gans}, and 
Kernel Inception Distance \cite{binkowski2018demystifying}) 
utilizes only images that were not {seen} in training.
We used 50K samples for metric evaluation.

For our generative adversarial network (GAN) baseline \cite{goodfellow2014generative}, we used a WGAN-GP generator for $64\times 64$ images, as implemented in the Mimicry library \cite{lee2020mimicry}, trained for 50K iterations. Hyper-parameter tuning was not necessary for strong performance.
For our variational autoencoder (VAE) baseline, we used the vanilla convolutional VAE architecture  \cite{kingma2013auto,rezende2014stochastic} from the PyTorch-VAE library \cite{Subramanian2020}. 
We tuned the learning rate and chose the $\beta$ value \cite{higgins2016beta} (the weight on the KL-divergence term) by iteratively decreasing it from the default value until the FID increased.
Note that both baselines were modified to handle the additional alpha channel (i.e., the mask) present in the data (and produced by the differentiable renderer in our main model).

For our unconditional generation procedure, based on \textit{ex-post} density fitting,
we use a Gaussian Mixture Model with 10 components, 
fit using the scikit-learn library \cite{scikit-learn}.
We use only the $v$ values inferred from the graphics cycle 
(i.e., only those coming from real \textit{shapes}, via $v = f_v(P)$).

\newcommand{\wwy}{0.317}
\begin{figure*}[h]
	\centering
	\includegraphics[width=\wwy\textwidth]{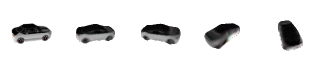} \hfill
	\includegraphics[width=\wwy\textwidth]{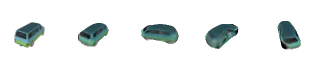} \hfill
	\includegraphics[width=\wwy\textwidth]{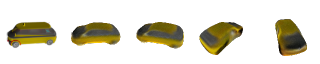} \\
	\includegraphics[width=\wwy\textwidth]{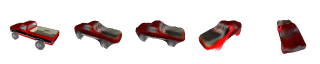} \hfill
	\includegraphics[width=\wwy\textwidth]{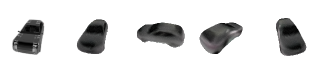} \hfill
	\includegraphics[width=\wwy\textwidth]{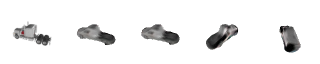} \\
	\includegraphics[width=\wwy\textwidth]{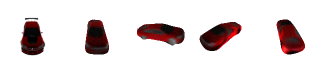} \hfill
	\includegraphics[width=\wwy\textwidth]{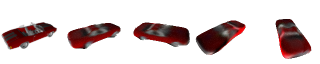} \hfill
	\includegraphics[width=\wwy\textwidth]{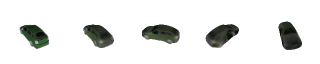} \\
	\includegraphics[width=\wwy\textwidth]{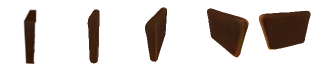} \hfill
	\includegraphics[width=\wwy\textwidth]{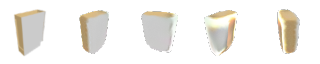} \hfill
	\includegraphics[width=\wwy\textwidth]{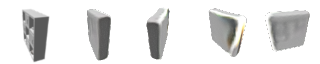} \\
	\includegraphics[width=\wwy\textwidth]{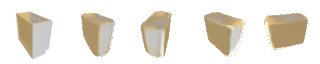} \hfill
	\includegraphics[width=\wwy\textwidth]{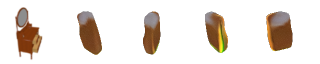} \hfill
	\includegraphics[width=\wwy\textwidth]{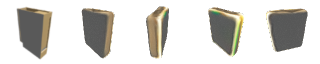} \\\includegraphics[width=\wwy\textwidth]{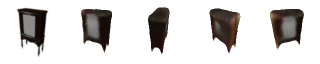} \hfill
	\includegraphics[width=\wwy\textwidth]{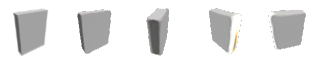} \hfill
	\includegraphics[width=\wwy\textwidth]{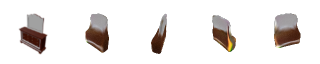} \\
	\includegraphics[width=\wwy\textwidth]{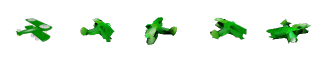} \hfill
	\includegraphics[width=\wwy\textwidth]{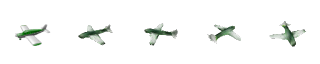} \hfill
	\includegraphics[width=\wwy\textwidth]{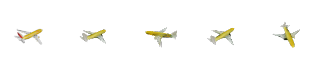} \\
	\includegraphics[width=\wwy\textwidth]{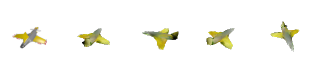} \hfill
	\includegraphics[width=\wwy\textwidth]{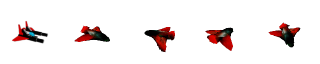} \hfill
	\includegraphics[width=\wwy\textwidth]{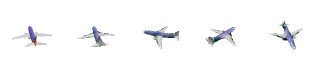} \\
	\includegraphics[width=\wwy\textwidth]{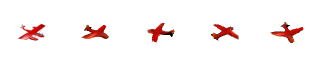} \hfill
	\includegraphics[width=\wwy\textwidth]{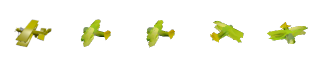} \hfill
	\includegraphics[width=\wwy\textwidth]{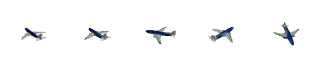} \\
	\includegraphics[width=\wwy\textwidth]{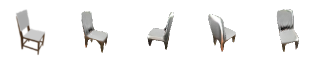} \hfill
	\includegraphics[width=\wwy\textwidth]{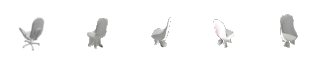} \hfill
	\includegraphics[width=\wwy\textwidth]{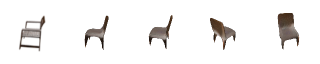} \\
	\includegraphics[width=\wwy\textwidth]{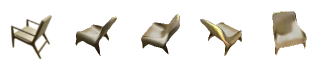} \hfill
	\includegraphics[width=\wwy\textwidth]{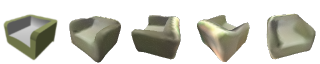} \hfill
	\includegraphics[width=\wwy\textwidth]{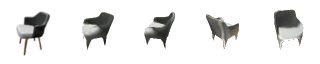} \\
	\includegraphics[width=\wwy\textwidth]{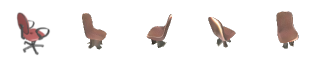} \hfill
	\includegraphics[width=\wwy\textwidth]{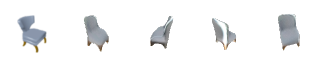} \hfill
	\includegraphics[width=\wwy\textwidth]{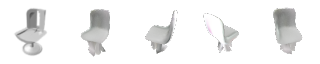} \\
	\includegraphics[width=\wwy\textwidth]{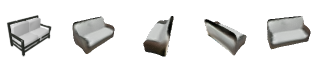} \hfill
	\includegraphics[width=\wwy\textwidth]{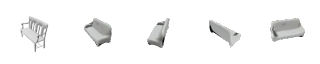} \hfill
	\includegraphics[width=\wwy\textwidth]{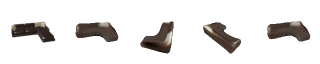} \\
	\includegraphics[width=\wwy\textwidth]{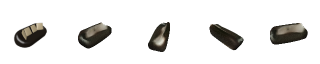} \hfill
	\includegraphics[width=\wwy\textwidth]{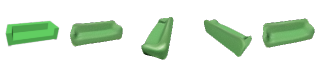} \hfill
	\includegraphics[width=\wwy\textwidth]{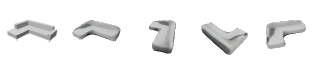} \\
	\includegraphics[width=\wwy\textwidth]{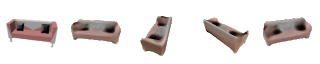} \hfill
	\includegraphics[width=\wwy\textwidth]{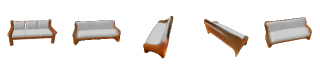} \hfill
	\includegraphics[width=\wwy\textwidth]{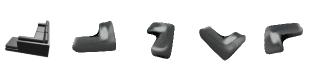} \\
	\caption{Example 3D inferences. Per inset (left to right): input $I$, reconstruction $\widehat{I}$, and different fixed views of the inferred textured mesh. }
	\label{fig:supp:recon}
\end{figure*}

\subsubsection{Ablation Experiments}

We also considered the effect of ablating (i) the pose-texture disentanglement loss (i.e., the objective maximizing the error of the adversary $A_R$) and (ii) the texture image critic (i.e., the term minimizing the error of $C_T$). We find the ablations lower the generative image quality: 
for the cars model,
in terms of IS, FID, and KID, we obtain
(i) 3.07, 131.2, and 11.7,
and
(ii) 2.84, 160.7, and 14.7, respectively,
all of which are worse than their non-ablated counterparts.

\newcommand{\xzz}{0.4758}
\begin{figure}
	\centering
	\includegraphics[width=\xzz\textwidth]{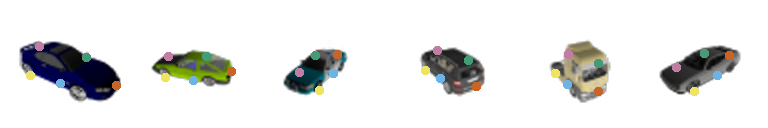}
	\includegraphics[width=\xzz\textwidth]{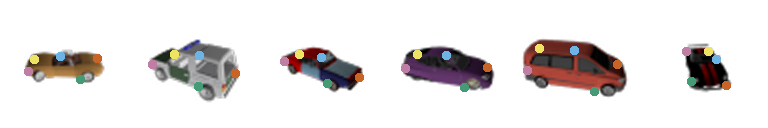}	
	\includegraphics[width=\xzz\textwidth]{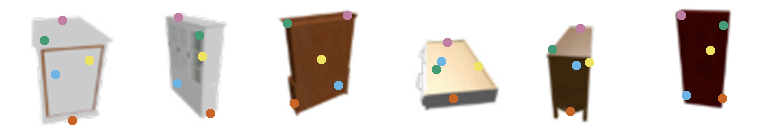}
	\includegraphics[width=\xzz\textwidth]{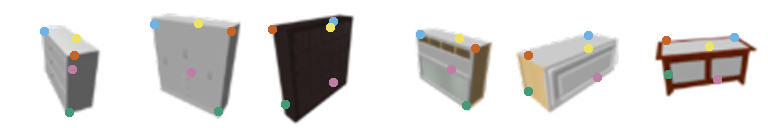}	
	\includegraphics[width=\xzz\textwidth]{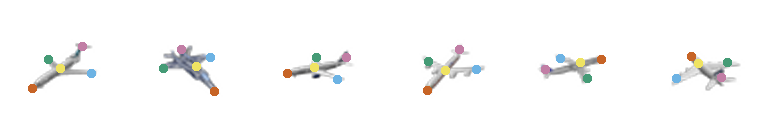}
	\includegraphics[width=\xzz\textwidth]{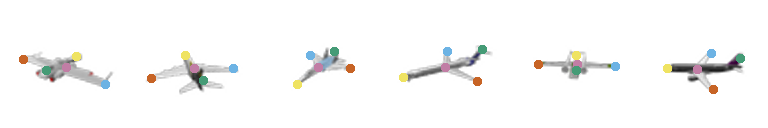}	
	\includegraphics[width=\xzz\textwidth]{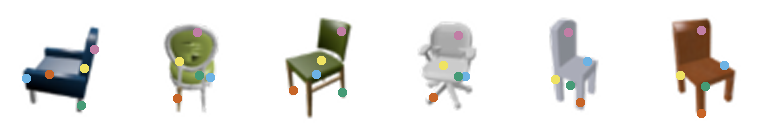}
	\includegraphics[width=\xzz\textwidth]{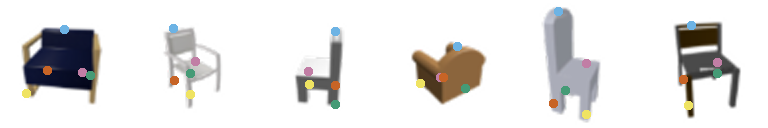}	
	\includegraphics[width=\xzz\textwidth]{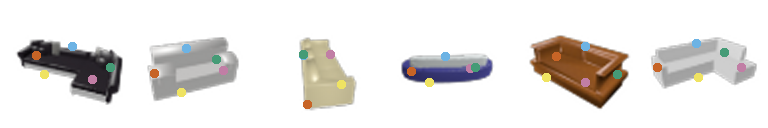}
	\includegraphics[width=\xzz\textwidth]{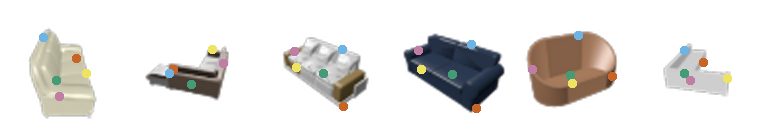}	
	\caption{Example unsupervised correspondence visualizations. 
		After inferring the template shape from each given input image $I$, we project the deformed and transformed template vertices of $\widetilde{M}_E$ back into the image $I$. 
		We highlight the positions of five randomly chosen vertices per image (but choose the same indices for each row).
		In other words, the chosen indices are the same for a single row (i.e., across columns), but different for every row (i.e., across rows), even for the same category.  }
	\label{fig:supp:realcorres}
\end{figure}

\newcommand{\hgq}{0.479}
\begin{figure}
	\centering
	\includegraphics[width=\hgq\textwidth]{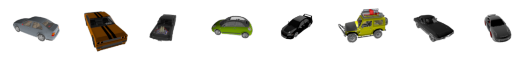} \\
	\includegraphics[width=\hgq\textwidth]{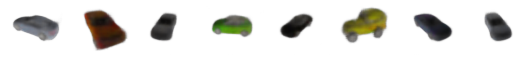} \\
	\includegraphics[width=\hgq\textwidth]{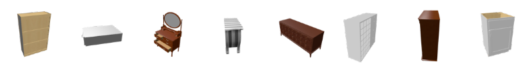} \\
	\includegraphics[width=\hgq\textwidth]{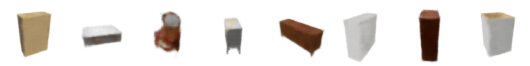} \\
	\includegraphics[width=\hgq\textwidth]{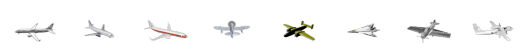} \\
	\includegraphics[width=\hgq\textwidth]{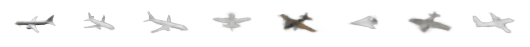} \\
	\includegraphics[width=\hgq\textwidth]{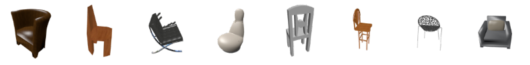} \\
	\includegraphics[width=\hgq\textwidth]{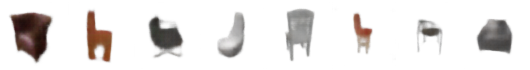} \\
	\includegraphics[width=\hgq\textwidth]{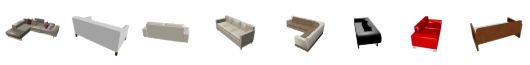} \\
	\includegraphics[width=\hgq\textwidth]{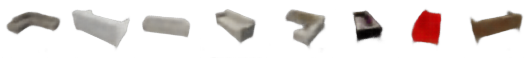} \\
	\caption{Example reconstructions from the VAE baseline model.}
	\label{fig:supp:vaerecon}
\end{figure}

\section{Additional Results Visualizations}
\label{suppsec:vis}

Lastly, we display additional visualizations of example results, in the same format as shown in the main text.
In Fig.\ \ref{fig:supp:recon}, we show
3D reconstruction examples (i.e., running the \textit{vision cycle} on an input image, to obtain an inferred textured mesh, which ``explains'' the original input image, and renders thereof).
In Fig.\ \ref{fig:supp:gen}, we present random conditional generations
(i.e., applying the shape-to-image modality translation to shape data samples), 
as well as real images for comparison.
Similarly, in Fig.\ \ref{fig:supp:uncond_gen}, we show random \textit{un}conditional generations 
(via sampling through our \textit{ex-post} fitted distribution to the latent shape, as described in the main text).
For additional comparisons to the baseline generative models (GAN and VAE),
we display random samples in Figs.\ \ref{fig:supp:baselinegen} for both
(we also show VAE reconstructions in Fig.\ \ref{fig:supp:vaerecon}).
In Fig.\ \ref{fig:supp:interps}, we visualize latent space interpolations,
illustrating (a) the disentanglement between shape, pose, and texture and (b) the smoothly manipulable nature of our latent representation.
Lastly, in Fig.\ \ref{fig:supp:realcorres}, we show our unsupervised correspondence visualizations. By marking the positions of fixed template vertices across input images, we see that the network learns to assign implicit meaning to a given node. A given template node can thus act like an unsupervised keypoint, which can identify correspondences across images.

\newcommand{\www}{0.92}
\begin{figure*}
	\centering
	\includegraphics[width=\www\textwidth]{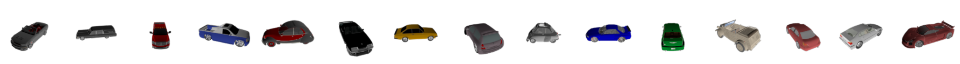}
	\includegraphics[width=\www\textwidth]{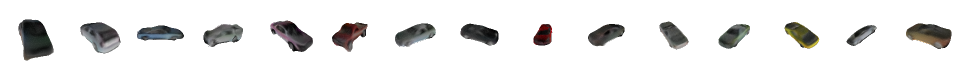}
	\includegraphics[width=\www\textwidth]{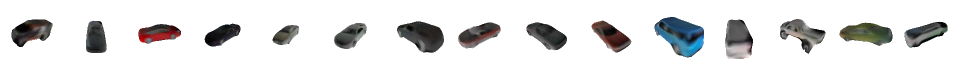}
	\includegraphics[width=\www\textwidth]{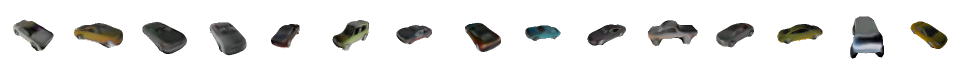}	
	\includegraphics[width=\www\textwidth]{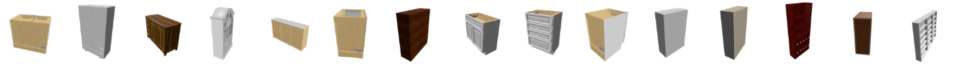}
	\includegraphics[width=\www\textwidth]{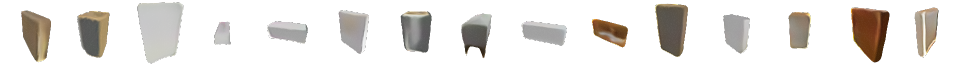}
	\includegraphics[width=\www\textwidth]{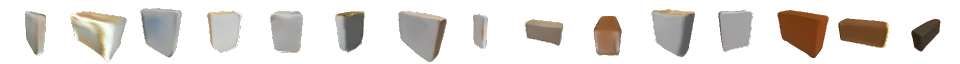}
	\includegraphics[width=\www\textwidth]{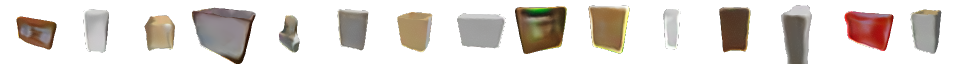}	
	\includegraphics[width=\www\textwidth]{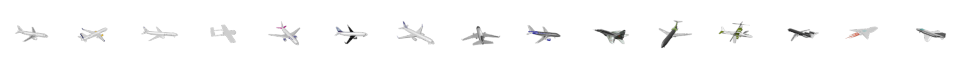}	
	\includegraphics[width=\www\textwidth]{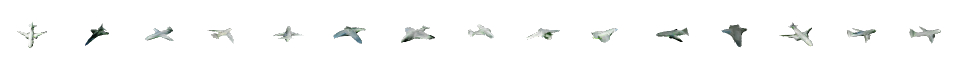}
	\includegraphics[width=\www\textwidth]{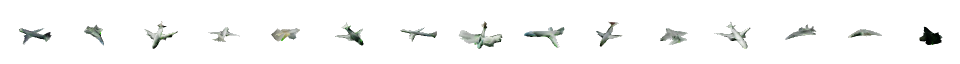}
	\includegraphics[width=\www\textwidth]{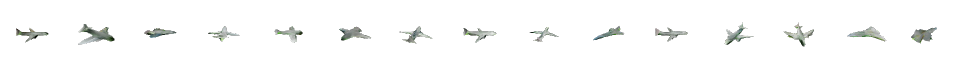}	
	\includegraphics[width=\www\textwidth]{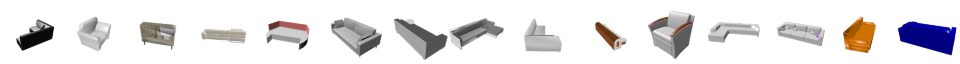}	
	\includegraphics[width=\www\textwidth]{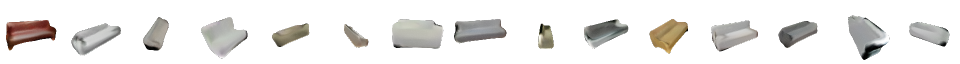}
	\includegraphics[width=\www\textwidth]{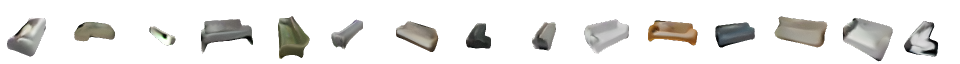}
	\includegraphics[width=\www\textwidth]{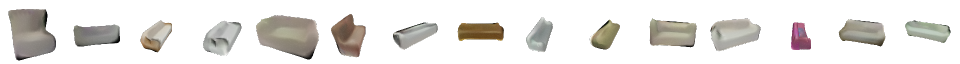}	
	\includegraphics[width=\www\textwidth]{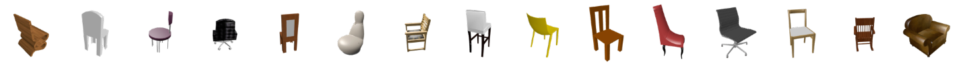}	
	\includegraphics[width=\www\textwidth]{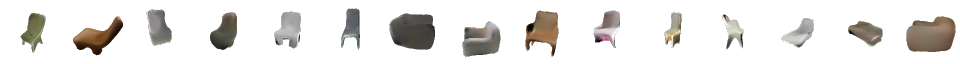}
	\includegraphics[width=\www\textwidth]{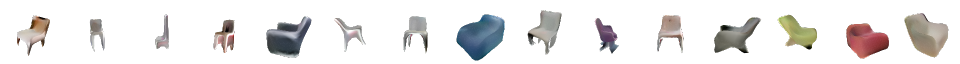}
	\includegraphics[width=\www\textwidth]{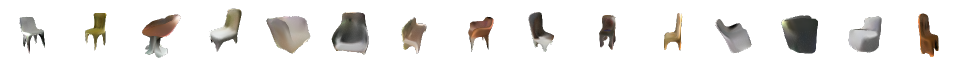}	
	\caption{Additional random sampled renders. Per category, we show one row of \textit{real} images, then three rows of generations. }
	\label{fig:supp:gen}
\end{figure*}

\newcommand{\wwwq}{0.925}
\begin{figure*}
	\centering
	\includegraphics[width=\wwwq\textwidth]{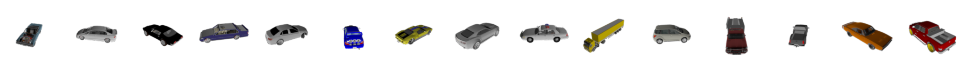}
	\includegraphics[width=\wwwq\textwidth]{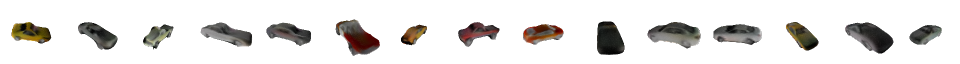}
	\includegraphics[width=\wwwq\textwidth]{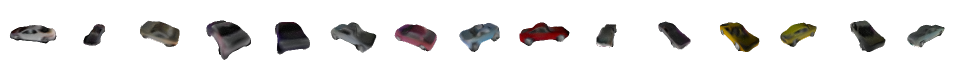}
	\includegraphics[width=\wwwq\textwidth]{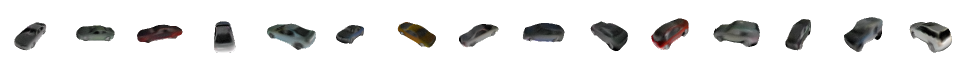}	
	\includegraphics[width=\wwwq\textwidth]{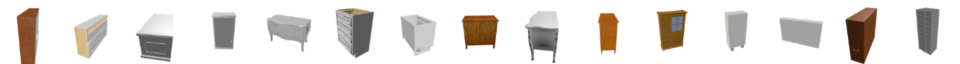}
	\includegraphics[width=\wwwq\textwidth]{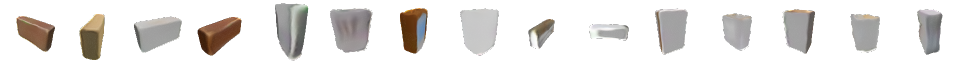}
	\includegraphics[width=\wwwq\textwidth]{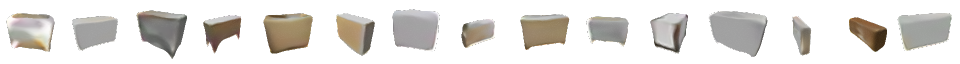}
	\includegraphics[width=\wwwq\textwidth]{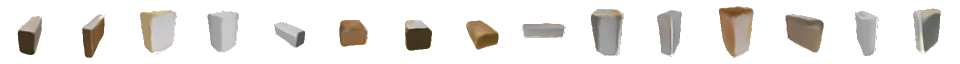}	
	\includegraphics[width=\wwwq\textwidth]{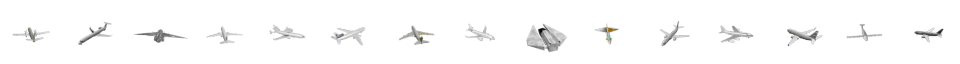}	
	\includegraphics[width=\wwwq\textwidth]{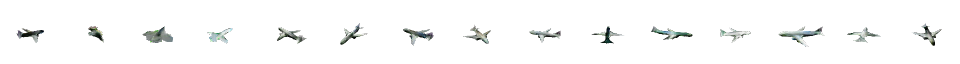}
	\includegraphics[width=\wwwq\textwidth]{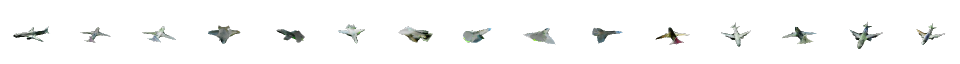}
	\includegraphics[width=\wwwq\textwidth]{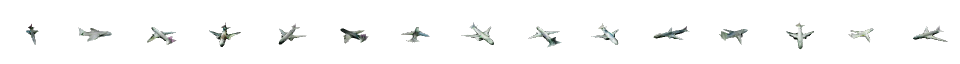}	
	\includegraphics[width=\wwwq\textwidth]{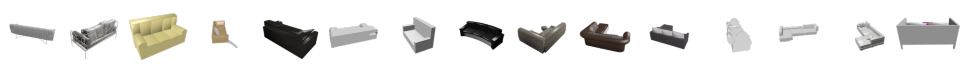}	
	\includegraphics[width=\wwwq\textwidth]{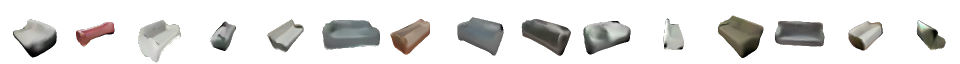}
	\includegraphics[width=\wwwq\textwidth]{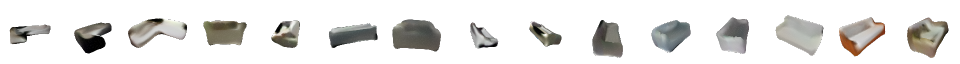}
	\includegraphics[width=\wwwq\textwidth]{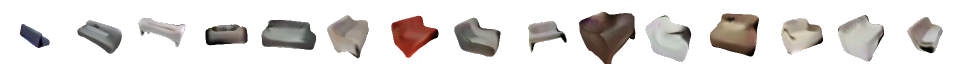}	
	\includegraphics[width=\wwwq\textwidth]{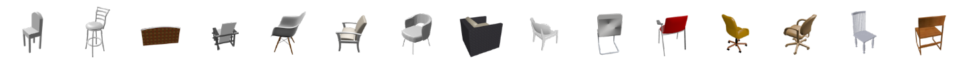}	
	\includegraphics[width=\wwwq\textwidth]{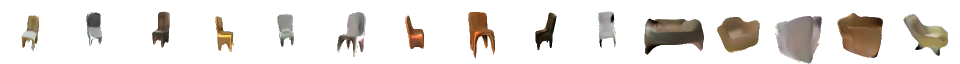}
	\includegraphics[width=\wwwq\textwidth]{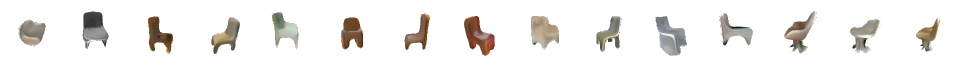}
	\includegraphics[width=\wwwq\textwidth]{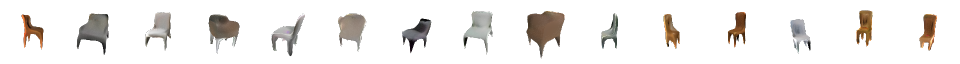}	
	\caption{\textit{Un}conditional generated render samples. 
		As in Fig.\ \ref{fig:supp:gen}, for each category, we show one row of real images, then three of samples. }
	\label{fig:supp:uncond_gen}
\end{figure*}

\newcommand{\wwz}{0.235}
\begin{figure*}
	\centering
	\includegraphics[width=\wwz\textwidth]{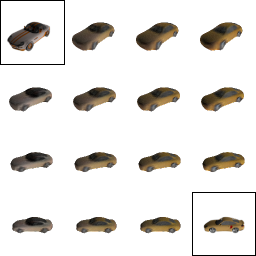} \hfill
	\includegraphics[width=\wwz\textwidth]{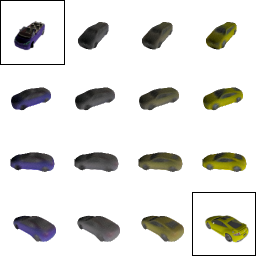} \hfill
	\includegraphics[width=\wwz\textwidth]{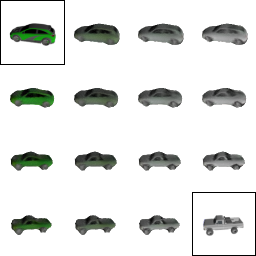} \hfill
	\includegraphics[width=\wwz\textwidth]{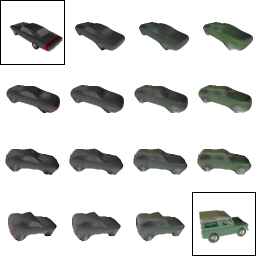} \\
	\includegraphics[width=\wwz\textwidth]{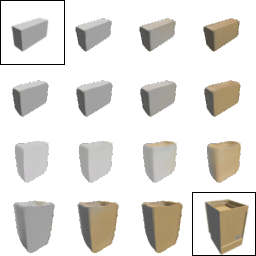} \hfill
	\includegraphics[width=\wwz\textwidth]{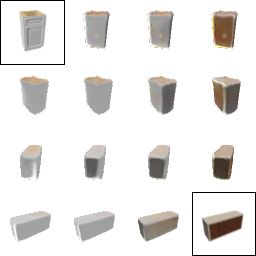} \hfill
	\includegraphics[width=\wwz\textwidth]{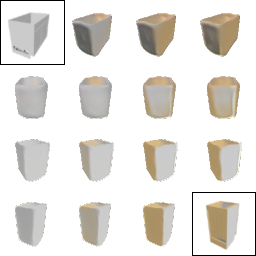} \hfill
	\includegraphics[width=\wwz\textwidth]{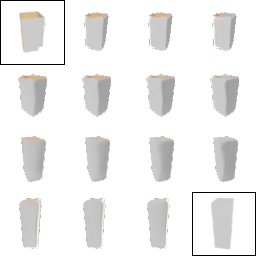} \\
	\includegraphics[width=\wwz\textwidth]{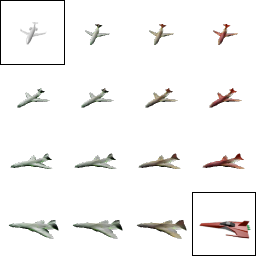} \hfill
	\includegraphics[width=\wwz\textwidth]{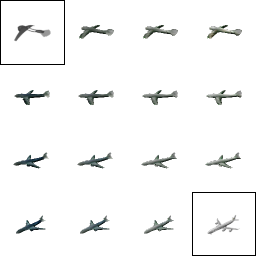} \hfill
	\includegraphics[width=\wwz\textwidth]{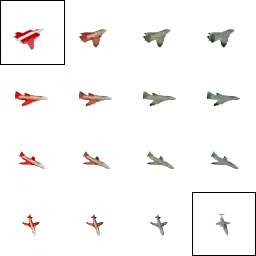} \hfill
	\includegraphics[width=\wwz\textwidth]{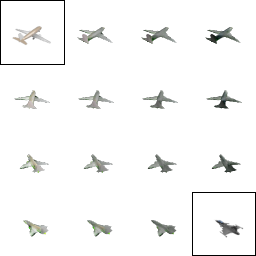} \\
	\includegraphics[width=\wwz\textwidth]{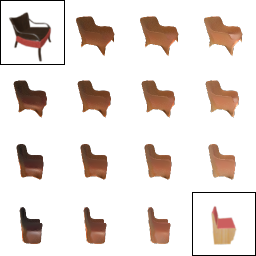} \hfill
	\includegraphics[width=\wwz\textwidth]{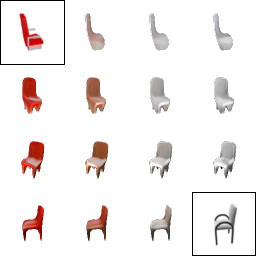} \hfill
	\includegraphics[width=\wwz\textwidth]{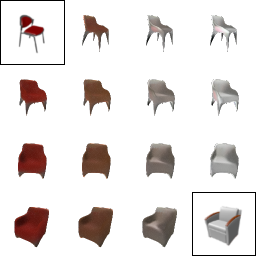} \hfill
	\includegraphics[width=\wwz\textwidth]{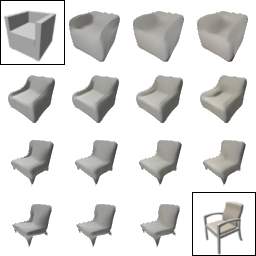} \\
	\includegraphics[width=\wwz\textwidth]{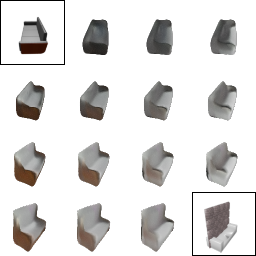} \hfill
	\includegraphics[width=\wwz\textwidth]{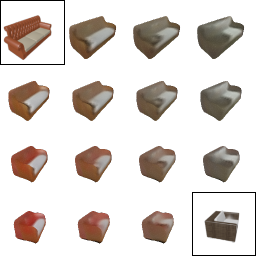} \hfill
	\includegraphics[width=\wwz\textwidth]{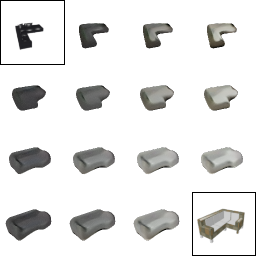} \hfill
	\includegraphics[width=\wwz\textwidth]{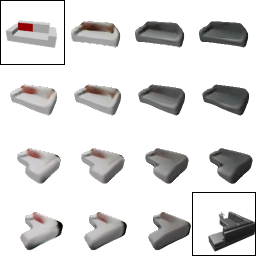} \\
	\caption{Example 3D latent interpolations. Per-inset: boxed corner images represent start and end points (real images), vertical axis corresponds to interpolation in latent shape ($v$) and Euclidean pose ($E$), while the horizontal axis represents interpolation of the latent texture ($\xi_T$). Non-boxed opposing corners (upper-right and lower-left) may be viewed as \textit{texture transfers}. Note the occasional ``flip'' of texture (rather than using the Euclidean transform $E$) to change orientation (e.g., cars, inset four, and planes, inset three). }
	\label{fig:supp:interps}
\end{figure*}

\newcommand{\hgy}{0.95}
\begin{figure*}
	\centering
	\includegraphics[width=\hgy\textwidth]{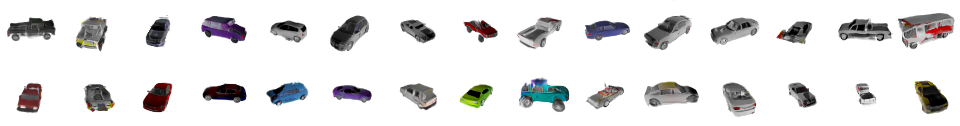} \\
	\includegraphics[width=\hgy\textwidth]{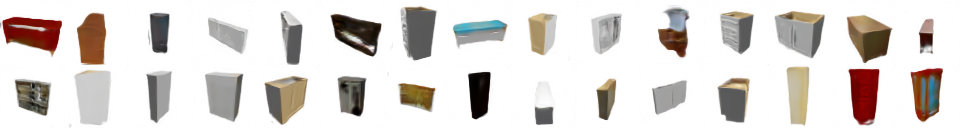} \\
	\includegraphics[width=\hgy\textwidth]{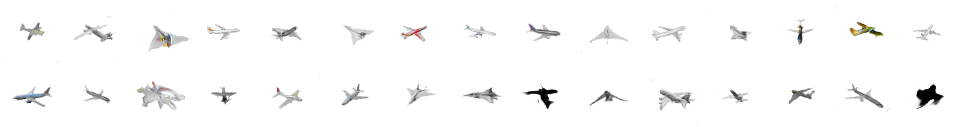} \\
	\includegraphics[width=\hgy\textwidth]{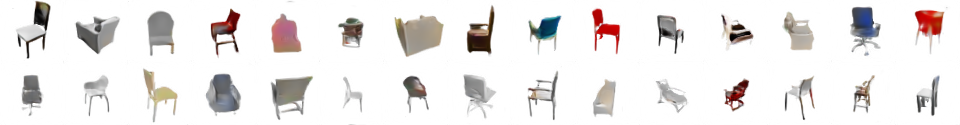} \\
	\includegraphics[width=\hgy\textwidth]{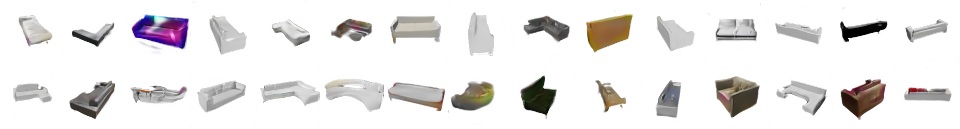} 	
	\includegraphics[width=\hgy\textwidth]{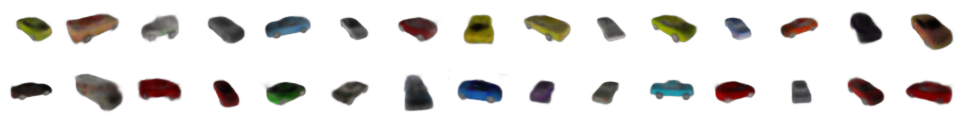} \\
	\includegraphics[width=\hgy\textwidth]{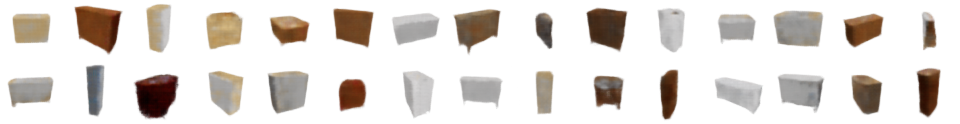} \\
	\includegraphics[width=\hgy\textwidth]{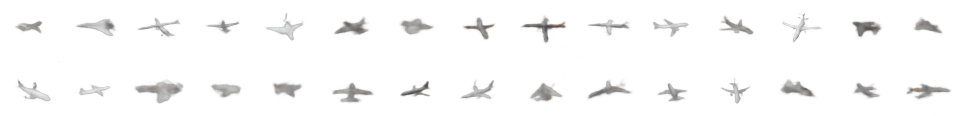} \\
	\includegraphics[width=\hgy\textwidth]{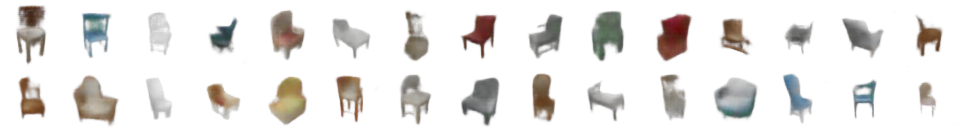} \\
	\includegraphics[width=\hgy\textwidth]{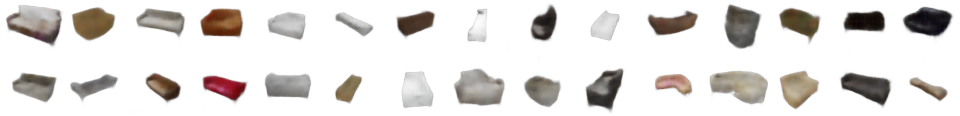} 	
	\caption{Example generations from the GAN (rows 1-10) and VAE (rows 11-20) baseline models.}
	\label{fig:supp:baselinegen}
\end{figure*}

\end{document}